# A Survey on Ensemble Learning under the Era of Deep Learning

Yongquan Yang[a], Haijun Lv[b], Ning Chen[c]

[a] Institution of Clinical Pathology, West China Hospital, Sichuan University, 37 Guo Xue Road, 610041 Chengdu, China

[b] AIDP, Baidu Co., Ltd, 701 Na Xian Road, 201210 Shanghai, China

[c] School of Electronics and Information, Xi'an Polytechnic University, 19 Jin Hua Road, 710048 Xi'an, China

**Abstract**

Due to the dominant position of deep learning (mostly deep neural networks) in various artificial intelligence applications, recently, ensemble learning based on deep neural networks (ensemble deep learning) has shown significant performances in improving the generalization of learning system. However, since modern deep neural networks usually have millions to billions of parameters, the time and space overheads for training multiple base deep learners and testing with the ensemble deep learner are far greater than that of traditional ensemble learning. Though several algorithms of fast ensemble deep learning have been proposed to promote the deployment of ensemble deep learning in some applications, further advances still need to be made for many applications in specific fields, where the developing time and computing resources are usually restricted or the data to be processed is of large dimensionality. An urgent problem needs to be solved is how to take the significant advantages of ensemble deep learning while reduce the required expenses so that many more applications in specific fields can benefit from it. For the alleviation of this problem, it is essential to know about how ensemble learning has developed under the era of deep learning. Thus, in this article, we present fundamental discussions focusing on data analyses of published works, methodologies, recent advances and unattainability of traditional ensemble learning and ensemble deep learning. We hope this article will be helpful to realize the intrinsic problems and technical challenges faced by future developments of ensemble learning under the era of deep learning.

Correspondence to: Yongquan Yang (remy_yang@foxmail.com)

# 1 Introduction

Ensemble learning (Dietterich 2000; Zhou 2009, 2012), a machine-learning technique that utilizes multiple base learners to form an ensemble learner for the achievement of better generalization of learning system, has achieved great success in various artificial intelligence applications and received extensive attention from the machine learning community. However, with the increase in the number of base learners, the costs for training multiple base learners and testing with the ensemble learner also increases rapidly, which inevitably hinders the universality of the usage of ensemble learning in many artificial intelligence applications. Especially when deep learning (LeCun et al. 2015) (mostly deep neural networks (He et al. 2016; Szegedy et al. 2017; Huang et al. 2017b; Xie et al. 2017; Sandler et al. 2018; Zoph et al. 2018; Tan and Le 2021)) is dominating the development of various artificial intelligence applications, the usage of ensemble learning based on deep neural networks (ensemble deep learning) is facing severe obstacles.

Since modern deep neural networks (Rawat and Wang 2017; Liu et al. 2017a; Alam et al. 2020; Khan et al. 2020) usually have millions to billions of parameters, the time and space overheads for training multiple base deep learners and testing with the ensemble deep learner is far greater than that of traditional ensemble learning. Fast ensemble deep learning algorithms like Snapshot (Huang et al. 2017a), fast geometric ensembling (FGE) (Garipov et al. 2018) and stochastic weight averaging (SWA) (Izmailov et al. 2018; Maddox et al. 2019) have promoted the deployment of ensemble deep learning in some artificial intelligence applications to some extent. However, due to the large expenses compared with traditional ensemble learning, the deployment of ensemble deep learning algorithms still needs further advances in some specific fields, where the developing time and computing resources are usually restricted (f. e. the field of robotics vision (Yang et al. 2018, 2019)) or the data to be processed is of large dimensionality (f. e. the field of histopathology whole slide image analysis (Yang et al. 2020c, b)). In these specific fields, it is sometimes still difficult to deploy the traditional ensemble learning algorithms, which makes the deployment of ensemble deep learning even more challenging. Thus, an urgent problem needs to be solved is how to make full usage of the significant advantages of ensemble deep learning while reducing the expenses required for both training and testing so that many more artificial intelligence applications in specific fields can benefit from it.

To alleviate this problem, it is essential to know about how ensemble learning has developed under the era of deep learning. In this article, we achieve this by presenting discussions with our observations of various existing ensemble learning algorithms and related applications. Different from existing review articles (Sagi and Rokach 2018; Dong et al. 2020) that mostly discussed about traditional ensemble learning, reviews (Cao et al. 2020) that particularly discussed ensemble deep learning in bioinformatics, or specific technical innovations with ensemble learning introduced, such as GAN-Ensembles (Durugkar et al. 2017; Ghosh et al. 2018; Han et al. 2021), this article aims to reveal the intrinsic problems and technical challenges of deploying ensemble learning under the era of deep learning to a wider range of specific fields, with more fundamental discussions about the developing routes of both traditional ensemble learning and ensemble deep learning. First, through data analysis of published works, we show the prosperity of ensemble learning and the gap between ensemble deep learning and traditional ensemble learning. Second, we discuss the methodologies, recent advances and unattainability of traditional ensemble learning by analyzing existing well-known approaches, applications and giving some of our observations. Third, we discuss the methodologies, recent advances and unattainability of ensemble deep learning with analyses and observations in the context of usual and fast solutions. Finally, we discuss the whole paper, pointing out some possible future research directions.

## 2 Data Analysis of Published Works

In this section, we present some data analysis of published works to show the prosperity of ensemble learning (EL) and the gap between ensemble deep learning (EDL) and traditional ensemble learning (TEL).

### 2.1 Prosperity of EL

The research history of EL can be traced back to 1990 (Hansen and Salamon 1990; Schapire 1990). Due to its significant advantages in improving the generalization of learning systems, the research on the theoretical algorithms and applications of EL has always been a research hotspot. Dietterich (Dietterich 1997), an authoritative scholar in the field of machine learning, once listed EL as the first of four research directions in the field of machine learning in AI Magazine. In the past 30 years, the research of EL has made remarkable progress. As shown in Figure 1, the number of papers related to the topic of 'ensemble learning' and published in the core set of Web of Science from 1990 to 2019 has been increasing steadily year by year, which reflects an exponentially growing prosperity of EL.

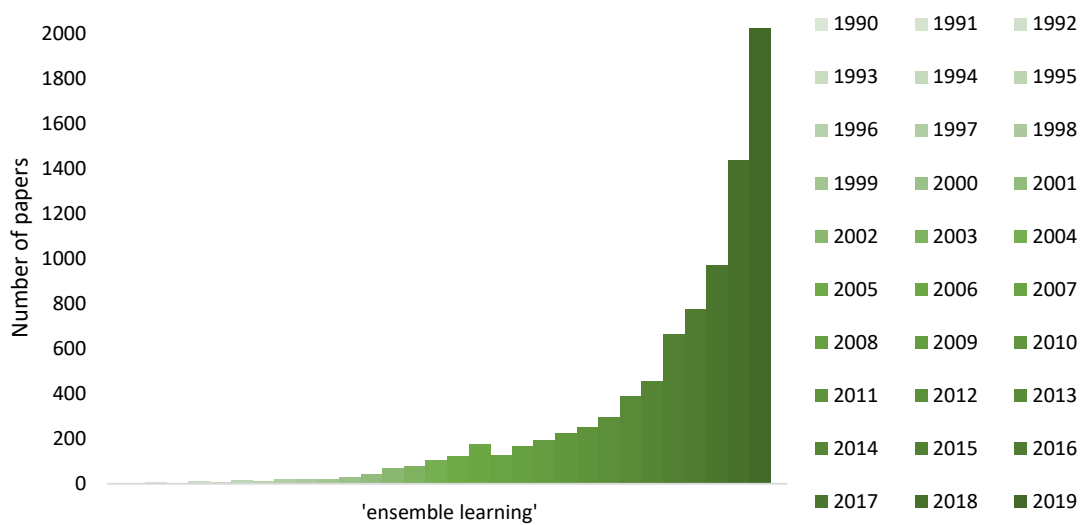

Figure. 1. The number of papers published in the core set of Web of Science from 1990 to 2019 for the topic of 'ensemble learning'.

## 2.2 Gap between EDL and TEL

The research history of ensembles of neural networks can be traced back to the early research stage of ensemble learning in 1990 (Hansen and Salamon 1990). However, the research progress of ensembles of deep neural networks (ensemble deep learning, EDL) had been very slow before 2014. Later benefited from the great success of deep neural networks with popularization of big data and high-performance computing resources, the research progress of EDL began to sprout rapidly. Figure 2 shows the comparison between the number of papers related to 'ensemble deep learning' or 'ensemble learning' not 'deep' (traditional ensemble learning, TEL) and published in the core set of Web of Science from 1990 to 2019. It can be noted from Figure 2 that a large gap exists between EDL and TEL. Compared with TEL, the development of EDL is severely lagging behind. The primary reason for this phenomenon is that modern deep neural networks (Rawat and Wang 2017; Liu et al. 2017a; Alam et al. 2020; Khan et al. 2020) usually have parameters ranging from millions to billions which makes the time and space overheads for the training and testing stages of EDL much greater than that of TEL, which severely hinders the deployment of EDL in various artificial intelligence applications in specifical fields.

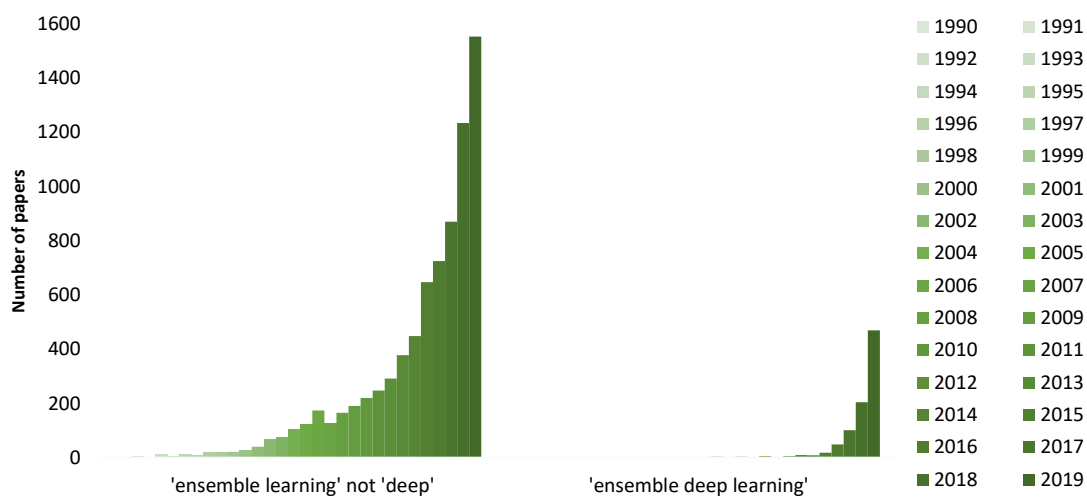

Figure 2. The comparison between the number of papers published in the core set of Web of Science from 1990 to 2019 for the topic of 'traditional ensemble learning' ('ensemble learning' not 'deep') and the topic of 'ensemble deep learning'.

## 3 Traditional Ensemble Learning

Traditional ensemble learning (TEL) has been playing a major role in the research history of ensemble learning (EL). In this section, starting with the paradigm of usual machine learning (UML), we respectively present the methodology of TEL, well-known implementations for the methodology of TEL, recent advances of TEL and unattainability of TEL.

### 3.1 Preliminary

Let us first consider the paradigm of usual machine learning (UML), where a raw data set $D = \{d_1, \cdots, d_n\}$ and its corresponding target set $T = \{t_1, \cdots, t_n\}$ are given. Specifically, $d_n$ is a raw data point of $D$ and $t_n$ is the target corresponding to $d_n$. In the UML paradigm, two components are normally essential: 1) feature extraction which converts raw data points into corresponding learnable representations; and 2) model development which evolves a learner that can map representations into corresponding targets.

The feature extraction consists of a number of extracting methods (Guyon and Elisseeff 2006), and a converting procedure. An extracting method is responsible for extracting a type of representations from raw data points; and the converting procedure is responsible for incorporating the results of extracting methods into learnable representations for the whole raw data set. The model development component consists of a learning algorithm, a learning strategy, and an evolving procedure. The learning algorithm is responsible for the construction and optimization of a learner; the learning strategy is responsible for configurating rules to carry out the evolving procedure; and the evolving procedure is responsible for, under the learning strategy, updating the parameters of the learner to be appropriate for mapping representations into corresponding targets. The outline for the methodology of UML is shown as Figure. 3.

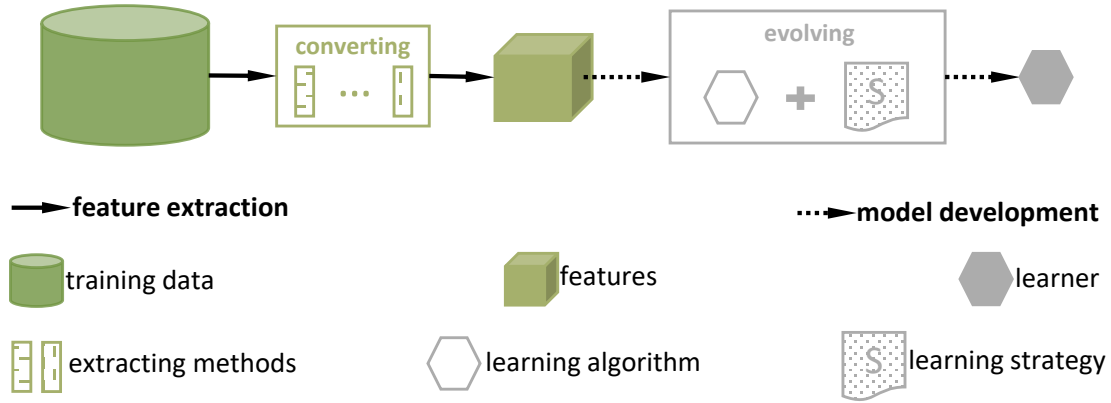

Figure. 3. The methodology of usual machine learning (UML), which consists of feature extraction and model development.

Formally, let $EM = \{em_1(*; \theta_1^{em}), \cdots, em_m(*; \theta_m^{em})\}$ denote various extracting methods. Particularly, $em_m(*; \theta_m^{em})$ signifies an extracting method $em_m(*)$ parameterized by $\theta_m^{em}$. The feature extraction component can be expressed as
$$F = FeatureExtraction(D, EM) = \{f_1, \cdots, f_n\}.$$
Thereinto, $f_n$, which signifies the representation of $d_n$, is constructed by the converting procedure that can be more specifically expressed as
$$f_n = converting(d_n, EM)$$
$$= \{em_1(d_n; \theta_1^{em}), \cdots, em_m(d_n; \theta_m^{em})\}$$
$$= \{x_{n,1}, \cdots, x_{n,k}\} \quad s.t. \ k \geq m.$$
Formally, let $LA = \{l(*; \theta^l), opt(*,*; \theta^{opt})\}$ denote the learning algorithm and

$LS = \{ls(\theta^{ls})\}$ denote the learning strategy. Specifically, $l(*; \theta^l)$ signifies the construction of a learner $l(*)$ parameterized by $\theta^l$ and $opt(*,*; \theta^{opt})$ signifies the learner's optimization procedure $opt(*,*)$ parameterized by $\theta^{opt}$ for updates of $\theta^l$, and $ls(\theta^{ls})$ signifies a learning strategy $ls$ parameterized by $\theta^{ls}$. Note, here $opt(*,*)$ implicitly consists of an objective function constructed for classification or regression and corresponding optimization of the objective function. The model development component can be expressed as

$$L = ModelDevelopment(F, T, LA, LS) = \{l(*; \theta_u^l)\}.$$

Thereinto, $l(*; \theta_u^l)$ signifies the learner $l(*)$ parameterized by $\theta_u^l$, which is updated by the evolving procedure that can be more specifically expressed as

$$\theta_u^l = evolving(F, T, l(*; \theta^l), opt(*,*; \theta^{opt}) \mid ls(\theta^{ls}))$$
$$= \arg \underset{\theta^l}{opt}\left(l(F; \theta^l), T; \theta^{opt} \mid ls(\theta^{ls})\right).$$

The developed learner $l(*; \theta_u^l)$ forms the mapping between representations $F$ and corresponding targets $T$.

At testing, given a test raw data point $d_{test}$, the corresponding target $t_{test}$ predicted by the evolved learner can be formally expressed as follows

$$f_{test} = \{em_1(d_{test}; \theta_1^{em}), \cdots, em_m(d_{test}; \theta_m^{em})\},$$
$$t_{test} = l(f_{test}; \theta_u^l).$$

Note, in the expressions of this subsection, each $\theta$ denotes the parameters corresponding to the implementation of respective expression.

### 3.2 Methodology of TEL

The paradigm of traditional ensemble learning (TEL) and the paradigm of UML primarily differ in the model development. TEL improves UML by replacing the model development of UML with generating base learners and forming ensemble learner. Generating base learners is responsible for evolving multiple base learners that have diversity in mapping the extracted representations into corresponding targets. Forming ensemble learner is responsible for integrating the base leaners into an ensemble leaner that can achieve better generalization.

With appropriate learning algorithms ($LA$) and learning strategies ($LS$), generating base learners can be roughly divided into two categories: one is to use different types of learning algorithms to generate 'heterogeneous' base learners; and the other is to use the same learning algorithm to generate 'homogeneous' base learners. Forming ensemble learner consists of an ensembling criteria and an integrating procedure. An ensembling criteria is responsible for the construction and configuration of an ensemble learner. With the ensembling criteria, an integrating procedure is responsible for forming the final ensemble learner which is appropriate for mapping from predictions of base learners into corresponding targets. The outline for the methodology of TEL is shown as Figure. 4.

Formally, generating base learners can be expressed as

$$L_b = BaseLearnerGeneration(F, T, LA, LS) = \{l_1(*; \theta_u^{l_1}), \cdots, l_b(*; \theta_u^{l_b})\}.$$

Thereinto, $l_b(*; \theta_u^{l_b})$ signifies a base learner $l_b(*)$ parameterized by $\theta_u^{l_b}$, which is updated by the evolving procedure. More specifically, the details of generating 'heterogeneous' base learners can be expressed as follows

$$LA = \left\{ \{l_1(*; \theta^{l_1}), opt_1(*,*; \theta^{opt_1})\}, \cdots, \{l_b(*; \theta^{l_b}), opt_b(*,*; \theta^{opt_b})\} \right\},$$

$$LS = \left\{ ls_{hetero,1}(\theta^{ls_{hetero,1}}), \cdots, ls_{hetero,b}(\theta^{ls_{hetero,b}}) \right\},$$

$$\theta_u^{l_b} = evolving\left(F, T, l_b(*; \theta^{l_b}), opt_b(*,*; \theta^{opt_b}) \mid ls_{hetero,b}(\theta^{ls_{hetero,b}})\right)$$

$$= \arg \underset{\theta^{l_b}}{opt} \left( l_b(F; \theta^{l_b}), T; \theta^{opt_b} \mid ls_{hetero,b}(\theta^{ls_{hetero,b}}) \right).$$

And, the details of generating 'homogeneous' base learners can be expressed as follows

$$LA = \left\{ \{l_0(*; \theta^{l_0}), opt_0(*,*; \theta^{opt_0})\} \right\},$$

$$LS = \left\{ ls_{homo,1}(\theta^{ls_{homo,1}}), \cdots, ls_{homo,b}(\theta^{ls_{homo,b}}) \right\},$$

$$l_b(*) = l_0(*),$$

$$\theta_u^{l_b} = evolving\left(F, T, l_0(*; \theta^{l_0}), opt_0(*,*; \theta^{opt_0}) \mid ls_{homo,b}(\theta^{ls_{homo,b}})\right)$$

$$= \arg \underset{\theta^{l_0}}{opt} \left( l_0(F; \theta^{l_0}), T; \theta^{opt_0} \mid ls_{homo,b}(\theta^{ls_{homo,b}}) \right).$$

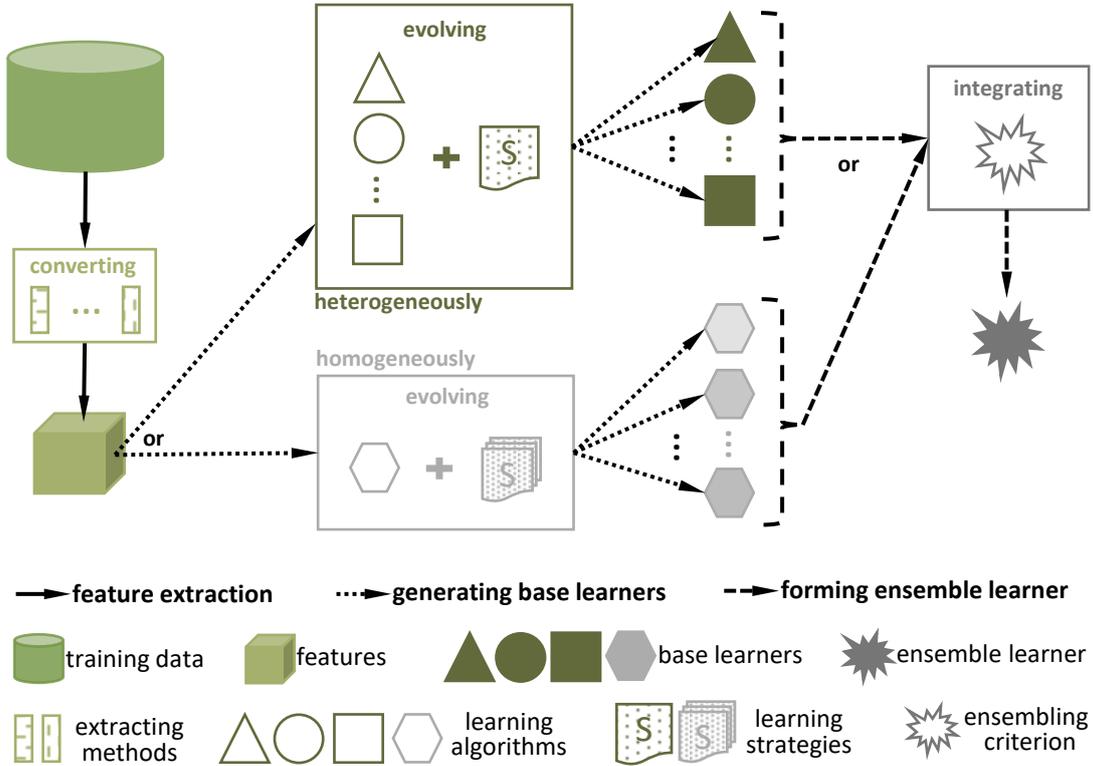

Figure. 4. The methodology of traditional ensemble learning (TEL), which consists of feature extraction, generating base learners and forming ensemble learner. In generating base learners, the learning strategy for evolving heterogeneous base learners is usually simpler than the learning strategy for evolving homogeneous base learners. In forming ensemble learner, building appropriate ensembling criteria is the key point to form the ensemble learner.

Let $EC = \{l_e(*; \theta^{l_e}), conf(*,*; \theta^{conf})\}$ denote the ensembling criteria for forming ensemble learner, and $\tilde{T} = \{\tilde{t}_1, \cdots, \tilde{t}_n\}, \tilde{t}_n = \{l_1(f_n; \theta_u^{l_1}), \cdots, l_b(f_n; \theta_u^{l_b})\}$ signify the predictions of the base learners. Specifically, $l_e(*; \theta^{l_e})$ signifies the construction of an ensemble learner $l_e(*)$ parameterized by $\theta^{l_e}$ and $conf(*,*$

; $\theta^{conf}$) signifies the ensemble learner's configuration $conf(*,*)$ parameterized by $\theta^{conf}$ for updates of $\theta^{l_e}$. Forming ensemble learning can be formally expressed as

$$L_e = EnsembleLearnerFormation(\tilde{T}, T, EC) = \{l_e(*; \theta_u^{l_e})\}.$$

Thereinto, $l_e(*; \theta_u^{l_e})$ signifies an ensemble learner $l_e(*)$ parameterized by $\theta_u^{l_e}$, which is configured by the integrating procedure. More specifically, the integrating procedure can be expressed as

$$\theta_u^{l_e} = integrating\left(\tilde{T}, T, l_e(*; \theta^{l_e}), conf(*,*; \theta^{conf})\right)$$
$$= arg\underset{\theta^{l_e}}{conf}\left(l_e(\tilde{T}; \theta^{l_e}), T; \theta^{conf}\right).$$

At testing, given a test raw data point $d_{test}$, the corresponding target $t_{test}$ predicted by the formed ensemble learner can be expressed formally as follows

$$f_{test} = \{em_1(d_{test}; \theta_1^{em}), \cdots, em_m(d_{test}; \theta_m^{em})\},$$
$$\tilde{t}_{test} = \{l_1(f_{test}; \theta_u^{l_1}), \cdots, l_b(f_{test}; \theta_u^{l_b})\},$$
$$t_{test} = l_e(\tilde{t}_{test}; \theta_u^{l_e}).$$

### 3.3 Well-known implementations for methodology of TEL

Since most existing well-known TEL approaches primarily focus on generating base learners and forming ensemble learner, here we weaken the descriptions for feature extraction (Guyon and Elisseeff 2006) since it is beyond the topic of the methodology of TEL. Particularly in discussion of generating base learners, we primarily elaborate on various learning strategies, as discussion of machine learning algorithms is a topic more appropriate in UML. The discussed well-known learning strategies and ensembling criteria for generating base learners and integrating base learners are summarized in Table. 1 and Table. 2.

Table. 1. Various well-known learning strategies (LS) for generating base learners

| Category | Specific LS | |
|---|---|---|
| Manipulating at data level | Sampling data points | Bagging (Breiman 1996) |
| | | Dagging (Ting and Witten 1997) |
| | | Boosting (Freund and Schapire 1997) |
| | Weighting data points | Wagging (Bauer and Kohavi 1999) |
| | | Multiboosting (Webb 2000) |
| Manipulating at feature level | Random Subspace (Ho 1998) | |
| Manipulating at both data and feature levels | Random Forests (Breiman 2001) | |
| | Rotation Forests (Rodríguez et al. 2006) | |
| Others | Negative correlation learning (Liu and Yao 1999) | |
| | ANN-based random initialization (Hansen and Salamon 1990) | |

Table. 2. Various well-known ensembling criteria (EC) for forming ensemble learner

| Category | Specific EC |
|---|---|
| Weighting methods | Majority voting / averaged summarization (Mendes-Moreira et al. 2012) |
| | Minimizing the objective (Mao et al. 2015) |
| Meta-learning methods | Stacking (Wolpert 1992) |

| | Mixture of expert (Masoudnia and Ebrahimpour 2014) |
| --- | --- |
| Ensemble selection (Zhou et al. 2002) methods (many could be better than all) | Weighting-based learning (Omari and Figueiras-Vidal 2015) |
| | Clustering based criteria (Bakker and Heskes 2003) |
| | Ranking-based criteria (Martinez-Muñoz et al. 2009) |
| | Selection based criteria (Dos Santos et al. 2008; Hernández-Lobato et al. 2009) |

**3.3.1 Base learner generation**

It is generally believed that the base learners used for ensemble should have sufficient diversity and the prediction of a single base learner should be as accurate as possible (Granitto et al. 2005; Zhou 2009, 2012). While generating multiple base learners, the learning algorithm and the learning strategy have to be considered. The learning algorithm used to generate the base learners can be a decision tree (DT) (Quinlan 1986), artificial neural network (ANN) (Rumelhart et al. 1986), support vector machine (SVM) (Chang and Lin 2011) or related variants. The used learning algorithm basically determines the upper limit of the predictive performance of the base learners. Generally, the predictive performance of the base learners is required to be at least better than random prediction, because, theoretically, a number of random predictions cannot be integrated to produce a well-regular prediction. The learning strategy that ensures the diversity of the base learners plays a decisive role in constructing the ensemble learner, especially when the accuracy of the base learners reaches certain limit. This is because, theoretically, a better ensemble learner cannot be summarized from a number of identical base learners. Usually, the learning strategy for generating base learners can be heterogeneous or homogeneous. When the learning strategy employs different learning algorithms to generate 'heterogeneous' base learners, the diversity of the base learners can be well ensured with very simple learning strategy like independently learning, because good diversity can exist among the base learners via the discrepancies of their diverse structures in addition to different parameters. However, when the learning strategy uses the same learning algorithm to generate 'homogeneous' base learners, the learning strategy to ensure the diversity of the base learners is the key point to success, because the diversity of the base learners is only kept in discrepancies of their parameters as they enjoy the same structure. Theoretically, both heterogeneous learning strategy and homogeneous learning strategy can be used at any time to generate base learners. However, the homogeneous learning strategy is the most commonly used, due to the fact that it only requires one learning algorithm to be implemented in practice. Thus, in the following contents of this subsection, we discuss various learning strategies for the diversity of 'homogeneous' base learners, respectively in the context of manipulating at data level, manipulating at feature level, manipulating at both data and feature levels, and others, which are summarized in Table 1.

(1) Manipulating at data level

Breiman et al. (Breiman 1996) and Freund et al. (Freund and Schapire 1997), ensured the diversity of the base learners by sampling or weighting the training set multiple times to generate new training sets for training different base learners. The two learning strategies are known as Bagging (Breiman 1996) and Boosting (Freund and Schapire 1997). Bagging uses the simple random sampling to generate new data sets for the training of base learners. A known variant of Bagging is Dagging (Ting and Witten 1997), which generates disjoint data sets by random sampling without replacement. Boosting starts with equal weights assigned to the samples of the training data and reweights the samples of the training data referring to the performance of previous base learner to focus on difficult data points. A known variant of Bagging is Wagging (Bauer and Kohavi 1999), which assigns random weights to the samples in the training data using Gaussian distribution or Poisson distribution. Combining Bagging and Wagging, Webb (Webb 2000) proposed Multiboosting, which firstly uses Boosting to assign weights to the samples in the training data, and then applies Wagging to update the weights of the assigned samples.

(2) Manipulating at feature level

Different from manipulating at the training data level (Breiman 1996; Freund and Schapire 1997; Ting and Witten 1997; Bauer and Kohavi 1999; Webb 2000), Ho (Ho 1998) upgraded the manipulation at data level to feature level (aka, Random Subspace) by sampling the features extracted from the training data and then using the sampled features as the input of the learning algorithm to generate various base learners.

(3) Manipulating at both data and feature levels

The two learning strategies of manipulating at data or feature level significantly promoted the development of ensemble learning. By combining the advantages of them, Breiman et al. (Breiman 2001) constructed a learning strategy by sampling the training data as well as the features extracted from the training data at the same time, and proposed the famous random forests (RF) ensemble learning approach, which is still widely used in various application fields. Later, Rodriguez et al. (Rodríguez et al. 2006) proposed rotation forests (RoF), the learning strategy of which first randomly splits the training features into several disjoint subsets and then principle component analysis (PCA) (Wold et al. 1987) is applied to each subset and axis rotation is employed to form new features for the training of base learners.

(4) Others

Another known learning strategy that can effectively ensure the diversity of base learners is to encourage negative correlation in the predictive error of base learners by complementary learning of the training data (Liu and Yao 1999). Besides, when artificial neural network (ANN) (Rumelhart et al. 1986) is employed as the learning algorithm, a commonly used learning strategy to obtain diverse base learners is training the ANN with different random initializations, which tends to differentiate the errors of the base learners (Hansen and Salamon 1990).

**3.3.2 Ensemble learner formation**

To form the ensemble learner from the based learners, an appropriate ensembling criterion is the key to success. Basically, known ensembling criteria for the ensemble learner formation can be roughly divided into three categories: weighting methods, meta-learning methods, and ensemble selection methods. In the following contents of this subsection, we respectively discuss these three categories, which are summarized in Table 2, and summarize corresponding usage.

(1) Weighting methods

The essence of the weighting methods for ensembling criteria is to combine the outputs of the base learners by assigning different weights to obtain better predictions. The majority voting rule commonly used for classification problem is the simplest implementation of the weighting method. The class with the most votes from the base learners will be taken as the final output of the ensemble learner. As for regression problem, the outputs of the base learners can be averaged (Mendes-Moreira et al. 2012) to produce the final output of the ensemble learner. Another commonly used weighting method is to assign the weight of a base learner in the ensemble learner according to its predictive ability on a validation data set. In addition, Mao et al. (Mao et al. 2015) used the error between the weighted output of multiple base learners and the true value as the objective function, where the weights of the base learners are subject to $\sum \omega_i = 1$ and $-1 < \omega_i < 1$, and minimized the objective function to determine the weights of the base learners for the ensemble learner. Weighting methods are simple to use, but they may not be able to fully exploit the information of the base learners, especially when large amount of data is available.

(2) Meta-learning methods

Unlike the weighting methods which simply assign different weights to the base learners, the fundamental idea of meta-learning methods for ensembling criteria is by learning from the meta-knowledge that has been learnt by the base learners (i.e., the predictions of the base learners) and corresponding targets to further reduce the generalization error. Via learning to map the predictions of the base learners into corresponding targets, meta-learning methods can further reduce the generalization error when large amount of data is available, by fully exploiting the information of the base learners to overcome the limitation of weighting methods. As the essence of meta-learning methods follows the paradigm of machine learning, they are appropriate to form ensemble learners for both classification and regression problems. In 1992, Wolpert (Wolpert 1992) proposed the first meta-learning method Stacking to integrate multiple base learners. A popular variant of the Stacking method is the mixture of experts (ME) (Masoudnia and Ebrahimpour 2014), which first uses a divide-and-conquer strategy to split the problem into multiple sub-problems, and respectively trains an expert (base learner) for each sub-problem, and then use another machine learning algorithm to learn from the outputs of the base learners and corresponding ground-truths to further reduce the generalization error. Combining the advantages of both the weighting method and the meta-learning method, Omari and FIgueiras-Vidal (Omari and Figueiras-Vidal 2015) first weighted the outputs of multiple base learners to obtain an initial ensemble, and then used another machine learning algorithm to learn from the initial ensemble to further reduce the generalization error. Meta-learning methods can fully exploit the information from the base learners, but it will take much more expenses to evolve the ensemble criteria, especially when the number of base learners is relatively large.

(3) Ensemble selection methods

In the early research stage of ensemble learning, most ensembling criteria used all the generated base learners to form an ensemble learner. Although this simple selection strategy can obtain an ensemble learner that is significantly better than a single base learner with appropriate ensembling criterion, the prediction time and storage space of the ensemble learner increased rapidly with the increase in the number of base learners. To alleviate this problem, in 2002, Zhou et al (Zhou et al. 2002) proposed the concept of ensemble selection for the first time, affirming that using a small number of base learners can as well achieve better ensemble performance. This caused a strong focus in the community of ensemble learning and opened a new research direction of selective ensemble learning (Zhang and Zhang 2011). The strategies for selecting the base learner can be roughly divided into: clustering-based methods, ranking-based methods, and selection-based methods. The clustering-based method needs to consider how to measure the similarity between the prediction results of two base learners (subsets), which clustering algorithm to choose, and how to determine the number of subsets for the base learners (Bakker and Heskes 2003). The ranking-based method first uses a certain measurement (such as accuracy) to rank the base learners, and then uses a suitable stopping criterion to select a number of base learners for ensemble (Martinez-Muñoz et al. 2009). The selection-based method leverages a certain selection strategy, such as greedy strategy (Jungnickel 1999) or dynamic programming (Kumar 2010), to select only a proportion of the base learners to participate in integrating (Dos Santos et al. 2008; Hernández-Lobato et al. 2009). Ensemble selection methods can favorably reduce the prediction time and storage space of the ensemble learner, but it may not achieve the optimal solution by ignoring the information of some base learners.

(4) Summarization of usage

In summary, there are no stationary rules for deciding which ensembling criterion is better than others. One has to construct the appropriate ensembling criterion by referring to specific situations. Theoretically, weighting methods can be used without any prerequisites. However, weighting methods may not be able to fully exploit the information of the base learners when large amount of collected data is available. In this situation, meta-learning methods are better than weighting methods. In addition, when the efficiency of the ensemble learner is a key factor to be considered, ensemble selection methods should also be united with weighting methods or meta-learning methods for a more appropriate ensembling criterion.

## 3.4 Recent advances of TEL

### 3.4.1 Material collection

Apart from the well-known implementation of TEL, in recent years, a lot of TEL advances have been proposed for various artificial intelligence applications (Behera and Mohanty 2019; Mohapatra et al. 2021). As the number of proposed TEL advances can reach up to hundreds or thousands in a single year, it is quite difficult for a survey to contain all of these advances. Thus, we propose to review a number of highly cited recent TEL advances and summarize corresponding key innovations to show some underlying trends of recent development of TEL. That highly cited recent advances are selected is based on one of the assumptions behind the PageRank (Page et al. 1998). In the case of this paper, that is the more citations a paper gets the more people agree with the approach presented in the paper. Thus, we consider summarizations from the review of the selection of highly cited recent TEL advances can more appropriately and stably reflect the primary underlying trends in recent development of TEL than the collection of all the recent TEL advances. The selected works for review of recent TEL advances were suggested by searching on Web of Science (accessed at February 3, 2021) with the filtering rules including being published in the last ten years and being highly cited or being a hotspot.

### 3.4.2 Review

Referring to the methodology of TEL presented in Figure. 4, but different from the discussion in Section 3.1 that focused more on learning strategies for generating base learners and ensembling criteria for forming ensemble learner, we add additional descriptions of extracting methods for feature representations as well as depictions of learning algorithms for generating base learners, since they also provide essential basis for TEL approaches to be carried out in specific applications.

For the 3D human action recognition, Wang et al. (Wang et al. 2014) proposed an ensemble learning approach called actionlet ensemble. Actionlet was defined as a conjunction of the features for a subset of the joints of the human skeleton. The 3D joint position feature (3D-JPF) and local occupancy patter (LOP) were extracted from the training images, and further transformed into temporal patterns for training via the proposed Fourier temporal pyramid (FTP) (Wang et al. 2014). Multiple actionlet classifiers based on an support vector machine (SVM) (Chang and Lin 2011) were trained with a proposed data mining strategy which aims to discover discriminative actionlet classifiers on the training dataset. A convex linear combination of the pre-trained multiple actionlet classifiers were proposed to form the final actionlet ensemble classifier.

For the prediction of protein-protein interactions, You et al. (You et al. 2013) employed auto covariance (AC) (Guo et al. 2008), conjoint triad (CT) (Shen et al. 2007), local descriptor (LD) (Davies et al. 2008) and Moran autocorrelation (MA) (Xia et al. 2009) to extract features from protein sequences. The extracted features were used to train multiple base learners based on a single hidden layer feed-forward neural networks, extreme learning machine (ELM) (Ding et al. 2015; Huang et al. 2015), with different random initializations. Majority voting was employed to integrate the trained multiple base learners into the final ensemble learner.

For the identification of DNase I hypersensitive sites, Liu et al. (Liu et al. 2016a) proposed an ensemble learning algorithm entitled iDHS-EL. Multiple base learners based on the random forests (RF) algorithm (Breiman 2001) were trained using three types of features including Kmer (Lee et al. 2011), reverse complement Kmer (RC-Kmer) (Gupta et al. 2008) and pseudo dinucleotide composition (PseDNC) (Chen et al. 2013) that reflect the characteristics of a DNA sequence from different angles. The final ensemble learner was formed by employing the grid search strategy to integrate the pre-trained multiple base learners. For a similar task of identifying N6-methyladenosine sites, Wei et al. (Wei et al. 2018) proposed an ensemble learning algorithm named M6APred-EL. Three feature representation algorithms, including PS(kmer)NP, PCPs (Liu et al. 2016b), and RFHC-GACs (Chen et al. 2016), were used to extract diverse features from RNA sequences. The extracted features were respectively employed to train SVM (Chang and Lin 2011) for multiple base predictors to identify N6-methyladenosine sites. The obtained multiple base predictors were integrated by majority voting to form the ensemble predictor.

For the landslide susceptibility mapping, Chen et al. (Chen et al. 2017) proposed several ensembles using multiple base learners that were heterogeneously trained with artificial neural network (ANN) (Rumelhart et al. 1986), maximum entropy (MaxEnt) (Phillips et al. 2004) and SVM (Chang and Lin 2011). Based on previous studies, eleven landslide conditioning factors such as elevation, slope degree, aspect, profile and plan curvatures, topographic wetness index (TWI), distance to roads, distance to rivers, normalized difference vegetation index (NDVI), land use, and lithology were selected for training. The proposed ensembles include ANN-SVM, ANN-MaxEnt, ANN-MaxEnt-SVM, and SVM-MaxEnt which were integrated by testing different basic mathematics (multiplication, division, addition and subtraction). For the same task, Pham et al. (Pham et al. 2017) assessed the application of the ensemble learning technique in the landslide problem, constructing several ensemble learners by combining the base classifier of multilayer perceptron (MLP) (Rosenblatt 1958) with Bagging (Breiman 1996), Dagging (Ting and Witten 1997), Boosting (Freund and Schapire 1997), Multiboosting (Webb 2000) and Rotation Forests (Rodríguez et al. 2006). The factors used for training were selected based on topography, physic-mechanical properties, locations and meteorology. The majority voting was employed to form the ensemble learner. In addition, Dou et al. (Dou et al. 2020) also constructed three types of ensemble learners for this task, combining SVM (Chang and Lin 2011) respectively with Bagging (Breiman 1996), Boosting (Freund and Schapire 1997) and Stacking (Wolpert 1992) ensembling strategies. The factors used for training were selected by referring to previous literatures, and the ensembling criteria employed to form the ensemble learner include majority voting and meta-learning.

For the short-term load forecasting, Li et al. (Li et al. 2020) proposed an ensemble learning algorithm based on wavelet transform (Daubechies and Bates 1993), extreme learning machine (ELM) (Ding et al. 2015; Huang et al. 2015) and partial least squares regression (PLSR) (Geladi and Kowalski 1986). Various wavelet transform specifications were used to generated different types of features from input load series. Each type of the generated features was individually leveraged to train multiple predictors based on extreme learning machine (ELM) (Ding et al. 2015; Huang et al. 2015). PLSR, which tackle the high degree of correlation between the individual forecast, was used to weight the multiple predictors to establish an accurate ensemble forecast.

For the multi-class imbalance learning, Bi et al. (Bi and Zhang 2018) proposed an ensemble learning algorithm named diversified error correcting output codes (DECOC) by introducing the error correcting output codes (ECOC) (Dietterich and Bakiri 1995) into the ensemble learning algorithm diversified one-against-one (DOVO) (Kang et al. 2015). ECOC is a decomposition strategy that builds a codeword for each class by maximizing the distance (such as Hamming distance) between various classes and, by selecting bits from the built codewords, divides the original problem into multiple sub-problems which can be further individually solved via machine learning. In original

DOVO, one-vs-one (OVO) was employed to divide the original problem into multiple sub-problems which are further solved with a variety of learning algorithms. OVO is a decomposition strategy that build a sub-problem by only selecting instances for each pair of classes from the original data. DECOC is formed by replacing OVO of EOVO with ECOC, and the diverse learning algorithms that DECOC employs include SVM (Chang and Lin 2011), k-nearest neighbor (KNN) (Altman 1992), Logistic Regression (LR) (Cox 1959), C4.5 (Salzberg 1994), AdaBoost (AB) (Freund and Schapire 1996), Random Forests (RF) (Breiman 2001), and multilayer perceptron (MLP) (Rosenblatt 1958). The obtained multiple based learners were weighted by minimizing the error in favor of the minority classes to form the final ensemble learner. For experiments, 17 public datasets were used, which had clearly defined features, the number of which ranges from 3 to 128.

For the soil moisture forecasting, Prasad et al. (Prasad et al. 2018) proposed and ensemble learning approach based on ensemble empirical mode decomposition (EEMD) (WU and HUANG 2009) and complete ensemble empirical mode decomposition with adaptive noise (CEEMDAN) (Torres et al. 2011). EEMD and CEEMDAN are two improved variants of empirical mode decomposition (EMD) (Huang et al. 1998), which decomposes intact soil moisture time series into several intrinsic mode functions (IMFs) to extract instantaneous frequency features. Each of the IMFs and residual were separately processed by partial autocorrelation function (PACF) (Ramsey 1974) to help selecting features with statistically significant relationship. The processed IMFs and residual were used to train extreme learning machine (ELM) (Ding et al. 2015; Huang et al. 2015) and Random Forests (RF) (Breiman 2001) to generate multiple base learners. To form the ensemble learner, the generated multiple base learners were integrated via averaged summation.

For the flood susceptibility mapping, Shahabi et al. (Shahabi et al. 2020) proposed an ensemble learning algorithm which employed four types of k-nearest neighbor (KNN) classifiers as the base learning algorithm and utilized Bagging (Breiman 1996) strategy to generate sub-datasets for training multiple base learners. Ten task-related conditioning factors (Rahmati et al. 2016) were chosen for training, including distance to river, elevation, slope, lithology, curvature, rainfall, topographic wetness index (TWI), stream power index (SPI), land use/land cover, and river density. The employed four types of KNN classifiers include coarse KNN (Altman 1992) which defines the nearest neighbor among all classes as the classifier, cosine KNN (CoKNN) (Hu et al. 2016) which uses the cosine distance metric as the nearest neighbor classifier, cubic KNN (CuKNN) (Wu et al. 2008) which uses the cubic distance metric as the nearest neighbor classifier, and weighted KNN which uses the weighted Euclidean distance as the nearest neighbor classifier. The obtained multiple base learners were integrated by majority voting to form the final ensemble learner.

For the automatic detection of lung cancer from biomedical dataset, Shakeel et al. (Shakeel et al. 2020) employed generalized neural network (GNN) (Behler and Parrinello 2007) as the learning algorithm and trained multiple base learners using the ensemble learning process proposed in (Xia et al. 2017), which generated multiple base learners with subsets equally divided from the input features collected by examining lung cancer effects on people's genetic changes and other impacts. The obtained multiple base learners were integrated by majority voting to form the final ensemble learner.

For the hyperspectral image classification, Su et al. (Su et al. 2020) proposed two types of ensemble learners which employed the tangent space collaborative representation classifier (TCRC) (Su et al. 2016) as the learning algorithm and respectively utilized Bagging (Breiman 1996) and Boosting (Freund and Schapire 1997) to train multiple base learners. The features used for training were the narrow spectral bands captured by hyperspectral images. TCRC is an improved collaborative representation classifier (CRC) (Li et al. 2016), the principle of which is that a testing sample can be approximated by training samples. CRC classifies a testing sample by assigning it to the class whose labeled training samples provide the smallest representation residual. TCRC improves CRC by using simplified tangent distance to take advantage of the local manifold in the tangent space of the testing sample (Su et al. 2016). The majority voting was employed to form the two types of ensemble learners.

For the wind power forecasting, Wang et al. (Wang et al. 2020) proposed a hybrid approach named BMA-EL based on Bayesian model averaging (BMA) (Hoeting et al. 1999; Wasserman 2000) and ensemble learning (EL). For training, the task-related features include wind speed, wind direction and ambient temperature. Artificial neural network (ANN) (Rumelhart et al. 1986), radial basis function neural network (RBFNN) (Borş and Pitas 1996), and SVM were employed as the learning algorithms. For generating more diverse base learners, the training set were divided into three subsets using clustering of self-organizing map (Vesanto and Alhoniemi 2000) and K-fold cross-validation (Bishop 2006). ANN, RBFNN and SVM were respectively optimized on the three generated subsets to produce three heterogeneous base learners. BMA was utilized to integrate the three base learners and the parameters of the BMA model were optimized on a validation dataset to form the final ensemble learner.

### 3.4.3 Summary

The information of the reviewed recent works of TEL is listed in Table. 3. The feature extraction methods adopted by the TEL approaches proposed in the reviewed works are listed in Table. 4. The learning algorithms and learning strategies employed to generate base learners are respectively listed in Table. 5 and Table. 6. The ensembling criteria utilized to integrate base learners are listed in Table. 7. Finally, referring to Table. 3-7, the key innovations of the TEL approaches proposed in the reviewed works are summarized in Table. 8.

Table. 3. The information of the applications in the reviewed recent TEL works (TA)

| App. | Description | | Citations | Published Year |
|---|---|---|---|---|
| TA1 | 3D human action recognition (Wang et al. 2014) | | 433 | 2014 |
| TA2 | Prediction of protein-protein interactions (You et al. 2013) | | 239 | 2013 |
| TA3 | Identification of hypersensitive sites | (Liu et al. 2016a) | 191 | 2016 |
| | | (Wei et al. 2018) | 91 | 2018 |
| TA4 | Landslide susceptibility mapping | (Chen et al. 2017) | 176 | 2017 |

|      |                                                         | (Pham et al. 2017) | 290 | 2017 |
|------|---------------------------------------------------------|--------------------|-----|------|
|      |                                                         | (Dou et al. 2020)  | 70  | 2020 |
| TA5  | Short-term load forecasting (Li et al. 2020)            |                    | 23  | 2020 |
| TA6  | Multi-class imbalance learning (Bi and Zhang 2018)      |                    | 97  | 2018 |
| TA7  | Soil moisture forecasting (Prasad et al. 2018)          |                    | 71  | 2018 |
| TA8  | Flood susceptibility mapping (Shahabi et al. 2020)      |                    | 45  | 2020 |
| TA9  | Detection of lung cancer (Shakeel et al. 2020)          |                    | 43  | 2020 |
| TA10 | Hyperspectral image classification (Su et al. 2020)     |                    | 26  | 2020 |
| TA11 | Wind power forecasting (Wang et al. 2020)               |                    | 24  | 2020 |

Table. 4. The methods for feature extraction (FE) of the TEL (TFE) approaches proposed in the reviewed works

| Code  | Description |
|-------|-------------|
| TFE1  | Temporal patterns transformed by Fourier temporal pyramid (FTP) (Wang et al. 2014) |
| TFE2  | Auto covariance (Guo et al. 2008) |
| TFE3  | Conjoint triad (Shen et al. 2007) |
| TFE4  | Local descriptor (Davies et al. 2008) |
| TFE5  | Moran autocorrelation (Xia et al. 2009) |
| TFE6  | Kmer (Lee et al. 2011) |
| TFE7  | Reverse complement Kmer (Gupta et al. 2008) |
| TFE8  | Pseudo dinucleotide composition (Chen et al. 2013) |
| TFE9  | PS(kmer)NP (Liu et al. 2016b) |
| TFE10 | PCPs (Liu et al. 2016b) |
| TFE11 | RFHC-GACs (Chen et al. 2016) |
| TFE12 | Landslide related factors (Chen et al. 2017) |
| TFE13 | Factors selected based on various properties of landslide (Pham et al. 2017) |
| TFE14 | Factors selected by referring to previous literatures (Dou et al. 2020) |
| TFE15 | Wavelet decompositions (Daubechies and Bates 1993) |
| TFE16 | Factors predefined on public datasets (Bi and Zhang 2018) |
| TFE17 | EEMD (WU and HUANG 2009) / CEEMDAN (Torres et al. 2011) followed with PACF (Ramsey 1974) |
| TFE18 | Flood related factors (Rahmati et al. 2016) |
| TFE19 | Factors of cancer effects (Shakeel et al. 2020) |
| TFE20 | Narrow spectral bands of hyperspectral images (Su et al. 2020) |
| TFE21 | Wind power related factors (Wang et al. 2020) |

Table. 5. The learning algorithms (LA) of the TEL (TLA) approaches proposed in the reviewed works

| Code | Abbr.  | Description |
|------|--------|-------------|
| TLA1 | SVM    | Support vector machine (Chang and Lin 2011) |
| TLA2 | ELM    | Extreme learning machine (Ding et al. 2015; Huang et al. 2015) |
| TLA3 | RF     | Random forest (Breiman 2001) |
| TLA4 | ANN    | Artificial neural network (Rumelhart et al. 1986) |
| TLA5 | MaxEnt | Maximum entropy (Phillips et al. 2004) |
| TLA6 | MLP    | Multilayer perceptron (Rosenblatt 1958) |
| TLA7 | KNN | K nearest neighbor (Altman 1992) |
| TLA8 | LR | Logistic regression (Cox 1959) |
| TLA9 | C4.5 | An improved iterative Dichotomiser 3 (Salzberg 1994) |
| TLA10 | AB | AdaBoost (Freund and Schapire 1996) |
| TLA11 | CoKNN | KNN based on cosine distance (Hu et al. 2016) |
| TLA12 | CuKNN | KNN based on cubic distance (Wu et al. 2008) |
| TLA13 | WeKNN | KNN based on weighted Euclidean distance |
| TLA14 | GNN | Generalized neural network (Behler and Parrinello 2007) |
| TLA15 | TCRC | Tangent space collaborative representation classifier (Su et al. 2016) |
| TLA16 | RBFNN | Radial basis function neural network (Borş and Pitas 1996) |

Table. 6. The learning strategies (LS) of the TEL approaches proposed in the reviewed works

| Code | Description |
|---|---|
| LS1 | Discovering discriminative learner on the training dataset (Wang et al. 2014) |
| LS2 | Training with randomly initialized parameters on the same dataset |
| LS3 | Manipulating at both the data and feature levels with random selections (Breiman 2001) |
| LS4 | Training multiple learners on the same dataset |
| LS5 | Manipulating at the feature level with division |
| LS6 | Manipulating at the data level with divide-and-conquer strategy |
| LS7 | Manipulating at the data level with Bagging strategy (Breiman 1996) |
| LS8 | Manipulating at the data level with Dagging strategy (Ting and Witten 1997) |
| LS9 | Manipulating at the data level with Boosting strategy (Freund and Schapire 1997) |
| LS10 | Manipulating at the data level with Multioosting strategy (Webb 2000) |
| LS11 | Manipulating at both the data and feature level with axis rotation (Rodríguez et al. 2006) |
| LS12 | Manipulating at the data level with self-organizing map (Vesanto and Alhoniemi 2000) and cross-validation (Bishop 2006) |

Table. 7. The ensembling criteria (EC) of the TEL (TEC) approaches proposed in the reviewed works

| Code | Description |
|---|---|
| TEC1 | Linear combination |
| TEC2 | Majority voting |
| TEC3 | Grid search |
| TEC4 | Basic mathematics |
| TEC5 | Stacking (Wolpert 1992) |
| TEC6 | Partial least squares regression (Geladi and Kowalski 1986) |
| TEC7 | Minimizing the error in favor of the minority classes |
| TEC8 | Averaged summation |
| TEC9 | Bayesian model averaging (Hoeting et al. 1999; Wasserman 2000) |

Table. 8. The key innovations of the TEL approaches proposed in the reviewed works

| App. | FE | LA&LS | | EC |
|---|---|---|---|---|
| | | LA | LS | |
| TA1 | TFE1 | TLA1 | Homo. by LS1 | W. by TEC1 |
| TA2 | TFE2-5 | TLA2 | Homo. by LS2 | W. by TEC2 |
| TA3 | TFE6-8 | TLA1 | Homo. by LS3 | W. by TEC3 |
| | TFE9-11 | TLA3 | Homo. by LS5 | W. by TEC2 |
| | TFE12 | TLA1,4,5 | Hetero. by LS4 | W. by TEC4 |
| TA4 | TFE13 | TLA6 | Homo. by LS7-10 | W. by TEC2 |
| | TFE14 | TLA7 | Homo. by LS7,9 | W. by TEC2 / M. by TEC5 |
| TA5 | TFE24 | TLA2 | Homo. by LS5 | W. by TEC6 |
| TA6 | TFE16 | TLA1,3,7-10 | Hetero. by LS6 | W. by TEC7 |
| TA7 | TFE17 | TLA2,3 | Homo. by LS5 | W. by TEC8 |
| TA8 | TFE18 | TLA7,11-13 | Hetero. / Homo. by LS7 | W. by TEC2 |
| TA9 | TFE19 | TLA14 | Homo. by LS5 | W. by TEC2 |
| TA10 | TFE20 | TLA15 | Homo. by LS7,9 | W. by TEC2 |
| TA11 | TFE21 | TLA1,4,16 | Hetero. by LS12 | W. by TEC9 |

Based on these tables, we can summarize: The representations of data employed by the proposed TEL approaches of the reviewed works for training mostly were various hand-crafted features; The learning algorithms of these TEL approaches that were frequently employed are classic SVM, KNN, and neural networks (ELM, ANN, MLP and GNN); Manipulating at the data level and the feature level by resampling were the frequently learning strategies used by these proposed TEL approaches; While the majority of these proposed TEL approaches trained homogenous (Homo.) base learners, three of them trained heterogenous (Hetero.) base learners; While the majority of these TEL approaches employed various weighting (W.) methods as ensembling criteria to form the final ensemble learner, only one of them tried meta-learning (M.) methods; And ensemble selection methods were rarely used, probably because the number of base learners was not massive enough to employ ensemble selection. These summarizations reflect the primary underlying trends for the recent development of TEL.

## 3.5 Unattainability

Based on the primary underlying trends of TEL summarized in the subsection 3.4.3, in this subsection, we discuss the unattainability of TEL. While most of recent advances of TEL focused on proposing solutions for specific applications based on combinatorial innovations by leveraging existing learning strategies for base learner generation and existing ensembling criteria for ensemble learner formation, few recent advances of TEL proposed new learning strategies for base learner generation or new ensembling criteria for ensemble learner formation. It seems that recent advances for the intrinsic problems of TEL have reached to a bottleneck, although TEL is still prosperously developing in various applications. Besides, the primary issue associated with the methodology of TEL comes from the nature of UML, which evolves a learner based on hand-crafted features which are usually difficult to design and not expressive enough.

# 4 Usual Ensemble Deep Learning

With the popularization of big data, computing resources, and deep learning (DL, mostly deep neural networks (He et al. 2016; Szegedy et al. 2017; Huang et al. 2017b; Xie et al. 2017; Sandler et al. 2018; Zoph et al. 2018; Tan and Le 2021)), which improves the paradigm of UML, has achieved unprecedented success in the field of machine learning. This will give birth to a huge number of approaches for ensembles of deep neural networks (ensemble deep learning, EDL) and promote the research of ensemble learning into a new era. The usual way to evolve the methodology of EDL (usual ensemble deep learning, UEDL) is directly applying DL to the methodology of TEL. In this section, starting with the paradigm of deep learning (DL), we respectively discuss the methodology, recent works and unattainability of UEDL.

## 4.1 Preliminary

Different from the paradigm of UML (Fig. 3), the paradigm of DL embeds the feature extraction into model development to form an end-to-end framework, which is able to learn task-specifically oriented features that are more expressive when massive training data is available. The paradigm of DL is shown as Fig. 5.

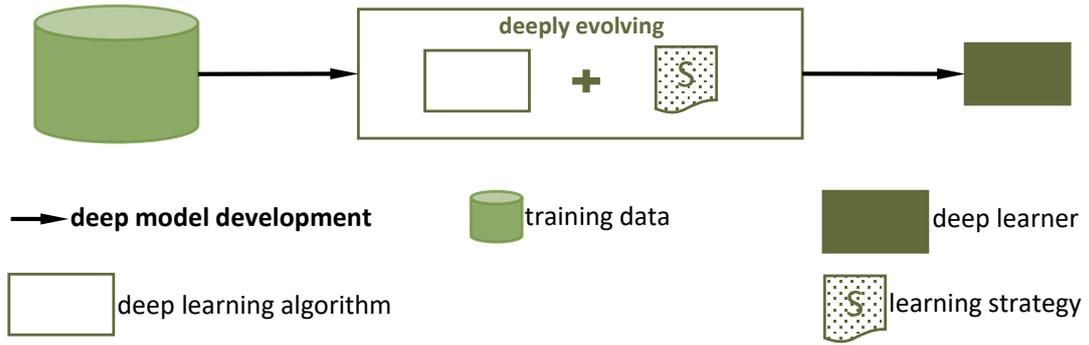

Figure. 5. The methodology of deep learning (DL), which embeds the feature extraction into deep model development to form an end-to-end framework.

Formally, let $DLA = \{dl(*; \theta^{dl}), dopt(*,*; \theta^{dopt})\}$ denote the deep learning algorithm and $LS = \{ls(\theta^{ls})\}$ denote the learning strategy. Specifically, $dl(*; \theta^{dl})$ signifies the construction of a deep learner $dl(*)$ parameterized by $\theta^{dl}$ and $dopt(*,*; \theta^{dopt})$ signifies the deep learner's optimization procedure $dopt(*,*)$ parameterized by $\theta^{dopt}$ for updates of $\theta^{dl}$. Note, here $dopt(*,*)$ implicitly consists of an objective function constructed for classification or regression and corresponding optimization of the objective function. The deep model development component can be expressed as
$$DL = DeepModelDevelopment(D, T, DLA, LS) = \{dl(*; \theta_u^{dl})\}.$$
Thereinto, $dl(*; \theta_u^{dl})$ signifies the deep learner $dl(*)$ parameterized by $\theta_u^{dl}$, which is updated by the deeply evolving procedure that can be more specifically expressed as
$$\theta_u^{dl} = deeply\_evolving(D, T, dl(*; \theta^{dl}), dopt(*,*; \theta^{dopt}) \mid ls(\theta^{ls}))$$
$$= \arg\underset{\theta^{dl}}{dopt}\Big(l(F; \theta^{dl}), T; \theta^{dopt} \mid ls(\theta^{ls})\Big).$$
The developed deep learner $dl(*; \theta_u^{dl})$ forms the mapping between data $D$ and corresponding targets $T$. Note the data $D$ here can be raw data or features extracted from raw data.

At testing, given a test data point $d_{test}$, the corresponding target $t_{test}$ predicted by the evolved deep learner can be expressed formally as follows
$$t_{test} = dl(d_{test}; \theta_u^{dl}).$$
Note, in the expressions of this subsection, each $\theta$ denotes the parameters corresponding to the implementation of respective expression.

## 4.2 Methodology of UEDL

Introducing DL to TEL, three basic patterns have emerged to evolve the methodology of UEDL. As shown in Fig. 6, the three basic patterns include: A) feature extraction based on DL; B) generating base learners based on DL; and C) forming ensemble learner based on DL.

Let $DEM = \{dem_1(*; \theta_1^{dem}), \cdots, dem_m(*; \theta_m^{dem})\}$ denote various DL-based extracting methods. Particularly, $dem_m(*; \theta_m^{dem})$ signifies a DL-based extracting method $dem_m(*)$ parameterized by $\theta_m^{dem}$. Specifically, $\theta_m^{dem} = \{dl_m(*; \theta_u^{dl_m}), \theta_m^{em}\}$, which signifies that $dem_m(*)$ consists of a developed deep learner $dl_m(*)$ parameterized by $\theta_u^{dl_m}$ and a hyperparameter $\theta_m^{em}$ indicating what features to extract from $dl_m(*; \theta_u^{dl_m})$. The pattern A can be formally expressed as
$$F = DeepFeatrueExtraction(D, DEM) = \{f_1, \cdots, f_n\}.$$
Thereinto, $f_n$, which signifies the deep representation of $d_n$, is constructed by a deeply converting procedure that can be more specifically expressed as
$$f_n = deeply\_converting(d_n, DEM)$$
$$= \{dem_1(d_n; \theta_1^{dem}), \cdots, dem_m(d_n; \theta_m^{dem})\}$$
$$= \{dx_{n,1}, \cdots, dx_{n,k}\} \quad s.t. \ k \geq m.$$

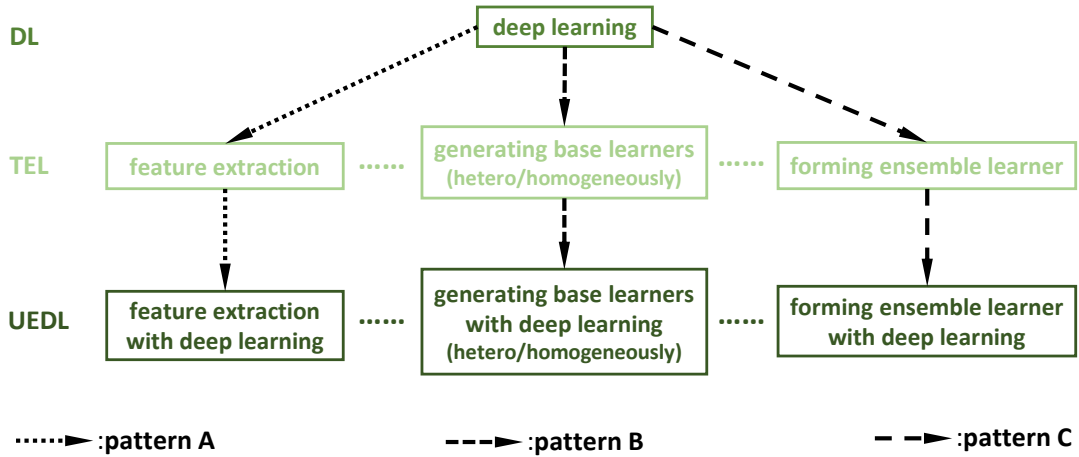

Figure. 6. Three basic patterns to evolve the methodology of usual ensemble deep learning (UEDL) by applying DL to TEL. Pattern A: feature extraction based on DL. Pattern B: generating base learners based on DL. Pattern C: forming ensemble learner based on DL.

With appropriate deep learning algorithms ($DLA$) and learning strategies ($LS$), the pattern B can be formally expressed as

$$L_b = BaseDeepLearnerGeneration(D,T,DLA,LS)$$
$$= \{dl_1(*;\theta_u^{dl_1}), \cdots, dl_b(*;\theta_u^{dl_b})\}.$$

Thereinto, $dl_b(*;\theta_u^{dl_b})$ signifies a base deep learner $dl_b(*)$ parameterized by $\theta_u^{dl_b}$, which is updated by the deeply evolving procedure. Identically, the data $D$ here can be raw data or features extracted from raw data. More specifically, the details of generating 'heterogeneous' base deep learners can be expressed as follows

$$DLA = \{\{dl_1(*;\theta^{dl_1}), dopt_1(*,*;\theta^{dopt_1})\}, \cdots, \{dl_b(*;\theta^{dl_b}), dopt_b(*,*;\theta^{dopt_b})\}\},$$
$$LS = \{ls_{hetero,1}(\theta^{ls_{hetero,1}}), \cdots, ls_{hetero,b}(\theta^{ls_{hetero,b}})\},$$
$$\theta_u^{dl_b} = deeply\_evolving\left(D, T, dl_b(*;\theta^{dl_b}), dopt_b(*,\right.$$
$$\left. *;\theta^{dopt_b}) \mid ls_{hetero,b}(\theta^{dl s_{hetero,b}})\right)$$
$$= \arg\,dopt_{\theta^{dl_b}}\left(dl_b(D;\theta^{dl_b}), T; \theta^{dopt_b} \mid ls_{hetero,b}(\theta^{ls_{hetero,b}})\right).$$

And, the details of generating 'homogeneous' deep base learners can be expressed as follows

$$DLA = \{\{dl_0(*;\theta^{dl_0}), dopt_0(*,*;\theta^{dopt_0})\}\},$$
$$LS = \{ls_{homo,1}(\theta^{ls_{homo,1}}), \cdots, ls_{homo,b}(\theta^{ls_{homo,b}})\},$$
$$dl_b(*) = dl_0(*)$$
$$\theta_u^{dl_b} = deeply\_evolving\left(D, T, dl_0(*;\theta^{dl_0}), dopt_0(*,\right.$$
$$\left. *;\theta^{dopt_0}) \mid ls_{homo,b}(\theta^{ls_{homo,b}})\right)$$
$$= \arg\,dopt_{\theta^{dl_0}}\left(dl_0(D;\theta^{dl_0}), T; \theta^{dopt_0} \mid ls_{homo,b}(\theta^{ls_{homo,b}})\right).$$

Let $DEC = \{dl_e(*;\theta^{dl_e}), dconf(*,*;\theta^{dconf})\}$ denote the ensemble criteria for ensemble deep learner formation, and $\tilde{T} = \{\tilde{t}_1, \cdots, \tilde{t}_n\}$ signify the predictions of the base learners. Specifically, $dl_e(*;\theta^{dl_e})$ signifies the construction of a DL-based ensemble learner $dl_e(*)$ parameterized by $\theta^{dl_e}$ and $dconf(*,*;\theta^{dconf})$ signifies the DL-based ensemble learner's configuration $dconf(*,*)$ parameterized by $\theta^{dconf}$ for updates of $\theta^{dl_e}$. The pattern C can be formally expressed as

$$L_e = EnsembleDeepLearnerFormation(\tilde{T},T,DEC) = \{dl_e(*;\theta_u^{dl_e})\}.$$

Thereinto, $dl_e(*;\theta_u^{dl_e})$ signifies an ensemble deep learner $dl_e(*)$ parameterized by $\theta_u^{dl_e}$, which is configured by the deeply integrating procedure. More specifically, the deeply integrating procedure can be expressed as

$$\theta_u^{dl_e} = deeply\_integrating\left(\tilde{T}, T, dl_e(*;\theta^{dl_e}), dconf(*,*;\theta^{dconf})\right)$$
$$= \arg\,dconf_{\theta^{dl_e}}(dl_e(\tilde{T};\theta^{dl_e}), T; \theta^{dconf}).$$

At testing, given a test data point $d_{test}$, the corresponding procedures for the three individual patterns can be expressed formally as follows

$$f_{test} = \{dem_1(d_{test};\theta_1^{dem}), \cdots, dem_m(d_{test};\theta_m^{dem})\},$$
$$\tilde{t}_{test} = \{dl_1(d_{test};\theta_u^{dl_1}), \cdots, dl_b(d_{test};\theta_u^{dl_b})\},$$
$$t_{test} = dl_e(\tilde{t}_{test};\theta_u^{dl_e}).$$

Note, the three patterns of introducing DL into TEL can be further combined to produce more patterns for the construction of UEDL. No matter which pattern is employed to construct an UEDL approach, the final formed learner for testing is usually regarded as an ensemble deep learner.

## 4.3 Recent advances of UEDL

### 4.3.1 Material collection

Though many recent advances (Panda et al. 2021; Mohammed and Kora 2021; Das et al. 2022) have been proposed for UEDL, we follow the strategy presented for the material collection of recent advances of TEL in section 3.4.1. Thus, we review a number of highly cited recent works that are related to the topic of UEDL and summarize corresponding key innovations to more appropriately and stably reflect the primary underlying trends in recent development of UEDL. Identically, the reviewed works were suggested by searching on Web of Science (accessed at March 1, 2021) with filtering rules including being published in last ten years, and being highly cited or being a hotspot.

### 4.3.2 Review

Referring to the methodology of TEL presented in Figure. 6, in addition to learning strategies for generating base learners and ensembling criteria for forming ensemble learner, we also add additional descriptions of extracting methods for feature representations as well as depictions of learning algorithms for generating base learners whenever they are necessary to provide essential basis for UEDL approaches to be carried out in specific applications.

For the probabilistic wind power forecasting, Wang et al. (Wang et al. 2017) proposed an ensemble deep learning approach based on different deep convolutional neural network (DCNN) (LeCun et al. 1998) architectures. Multiple base deep learners are independently trained using features extracted from raw wind power data by Mallat wavelet decomposition (Mallat 1989) and converted into 2D image. The employed DCNN architectures were different in terms of the number of hidden layers, the number of maps in each layer and the size of the input. For the training of each base deep learner, separated wavelet features were used. To form the ensemble deep learner, the predictions of the multiple base deep learners were fused by wavelet reconstruction (Daubechies and Bates 1993).

For the electricity load demand forecasting, Qiu et al. (Qiu et al. 2017) proposed a deep ensemble learning approach based on empirical mode decomposition (EMD) (Huang et al. 1998). First, via EMD, load demand time series were decomposed into several intrinsic mode functions (IMFs) and one residual which can be regarded as instantaneous frequency features extracted from the non-stationary and nonlinear raw data. Each of the IMFs and residual was separately used to train a deep belief network (DBN) (Hinton et al. 2006) which was composed of restricted Boltzmann machines (RBMs) (Hinton and Salakhutdinov 2006; Hinton 2012) and multilayer perceptron (MLP) (Gardner and Dorling 1998), resulting in a series of base deep learners. The base deep learners were integrated via averaged summation or weighted linear combination to form the ensemble learner.

For the sentiment analysis in social applications, Araque et al. (Araque et al. 2017) trained multiple base deep learners based on word2vec (Mikolov et al. 2013) / doc2vec (Le and Mikolov 2014) combined with logistic regression (word2vec/doc2vec-LR).

Word2vec was used for representations of short texts and doc2vec was used for representations of long texts. The obtained base deep learners were integrated to form the final ensemble deep learner through weighting by majority voting or meta-learning by random forests (Da Silva et al. 2014; Zhang and He 2015).

For the melanoma recognition in dermoscopic images, Codella et al. (Codella et al. 2017) employed convolutional architecture for fast feature embedding (Caffe) (Jia et al. 2014) and deep residual network (ResNet) (He et al. 2016) pretrained on ImageNet (Deng et al. 2009) and convolutional network of U-shape (U-Net) (Ronneberger et al. 2015) pretrained on lesion segmentation task (Codella et al. 2018) for feature representations, in addition to several types of traditional feature representations including color histogram (CH) (Barata et al. 2014), edge histogram (EH) (Barata et al. 2014), multi-scale variant of color local binary patters (MSLBP) (Zhu et al. 2010) and a type of sparse coding (SC) (Mairal 2014). The deep CNN based feature representations and the traditional feature representations were respectively employed to train SVM (Chang and Lin 2011) for multiple base learners. The predictions of the obtained base learners were averaged to form the final ensemble deep learner.

For the wind speed forecasting, Chen et al. (Chen et al. 2018) utilized long short-term memory (LSTM) (Hochreiter and Schmidhuber 1997; Sainath et al. 2015) to exploit the implicit information of collected wind speed time series. A cluster of LSTM with diverse number of hidden layers and neurons in each hidden layer were built to train multiple base deep learners on the collected raw data. A meta-learning strategy was proposed to form the final ensemble deep learner. The meta-learning strategy was composed of a support vector regression machine (SVRM) (Xu et al. 2017; Zhang et al. 2017) that took the forecasting results of the trained base deep learners as inputs and was optimized by extremal optimization (EO) (Boettcher and Percus 2000, 2001, 2002).

For the crude oil price forecasting, Li et al. (Zhao et al. 2017) formulated an ensemble deep learning approach named stacked denoising autoencoder bagging (SDAE-B). Employing the bagging strategy (Breiman 1996), multiple base deep learners were trained based on SDAE (Vincent et al. 2008, 2010). Referring to (Zagaglia 2010; Naser 2016), 198 factors, including price, flow, stock macroeconomic and financial series, were selected as exogenous variables for training. The predictions of the obtained base learners were averaged to form the final ensemble deep learner.

For the cancer prediction based on gene expression, Xiao et al. (Xiao et al. 2018) proposed to integrate multiple base learners using deep learning. The DEseq (Anders and Huber 2010) was employed to select informative genes from a given sequence for training. K nearest neighbor (KNN) (Altman 1992), support vector machine (SVM) (Chang and Lin 2011), decision tree (DT) (Quinlan 1986), random forests (RF) (Breiman 2001), and gradient boosting decision tree (GBDT) (Friedman 2001) were employed to train five base learners with model selection using K-fold cross-validation (Bishop 2006). To form the ensemble learner, a five-layer neural network (5-NN) was built as an implementation of the meta-learning strategy, which took the predictions of the five obtained base learners as inputs and output normal or tumor.

For the prediction of melt index (MI) in industrial polymerization processes, Liu et al. (Liu et al. 2018) proposed an ensemble deep kernel learning (EDKL) approach. A total of 11 factors correlated with MI were chosen for training, including the reactor pressure, reactor temperature, liquid level, and flow rate of the main catalysts. The deep belief network (DBN) (Hinton et al. 2006) comprising a series of individual restricted Boltzmann machines (RBM) (Hinton and Salakhutdinov 2006; Hinton 2012) was adopted to extracted features from the chosen factors in a unsupervised manner. With the extracted features, multiple base learners were trained based on a kernel learning soft-sensing model (KLSM) (Liu et al. 2013) by bagging (Breiman 1996). The predictions of the obtained base learners were averaged to form the final ensemble deep learner.

For the brain disease diagnosis based on MRI image, Suk et al. (Suk et al. 2017) as well proposed to integrate multiple base learners using deep learning. A series of techniques (details can be found in the preprocessing section) were employed to extract 93-dimensional volumetric features from an MRI image. Based on the extracted features, sparse regression model (SRM) (Wang et al. 2011; Zhang and Shen 2012) was utilized to train multiple base learners with different values of the regularization control parameter. To form the ensemble learner, a six-layer convolutional neural network (6-CNN) was built as an implementation of the meta-learning strategy, which took the predictions of the obtained base learners as inputs and output disease diagnosis.

For the red lesion detection in fundus images, Orlando et al. (Orlando et al. 2018) formed an ensemble learning approach enhanced by deep learning. An unsupervised candidate lesion detection approach was proposed based on morphological operations to identify potential lesion areas. A five-layer convolutional neural network (5-CNN) was built to learn features from patches around the detected potential lesion areas. A 63-dimensional feature vector was as well extracted from lesion candidates using descriptors explored in related literature (Niemeijer et al. 2005; SB and Singh 2012; Seoud et al. 2016). Both convolutional and hand-crafted features are used by random forests (RF) (Breiman 2001) to form the ensemble learner.

For bearing fault diagnosis, Xu et al. (Xu et al. 2019) prosed an ensemble learning approach based on deep learning and random forests (RF) (Breiman 2001). Morlet, a type of continuous wavelet transform, was employed to convert raw vibration signals of bearing, which was one-dimensional in time-domain, into spectrum of abundant condition information, which was two-dimensional in time-frequency domain (LIN and QU 2000). A built five-layer CNN model, which consists of two convolutional layers, two pooling layers and one fully connected layer, was trained with the converted spectrums for feature extraction. The features of the two pooling layers and the fully connected layer were independently employed by RF (Breiman 2001) to produce three RF base learners. To form the ensemble learner, the outputs of the three base learners were integrated by the winner-take-all strategy (Guo et al. 2016). Ma et al. (Ma and Chu 2019) also proposed an ensemble learning approach for the same task. Employing a multi-objective evolutionary algorithm (MOEA/DD) (Li et al. 2015), multiple base deep learners were generated based on three deep learning architectures, including DBN (Hinton et al. 2006), ResNet (He et al. 2016) and deep auto-encoder (DAE) (Jia et al. 2016). The three architectures all have five layers (including three hidden layers).

Regarding to the number of hidden neurons in each hidden layer and the learning rate for back propogation, MOEA/DD was leveraged to evolve a population of base deep learners with maximum accuracy and diversity simultaneously. Frequency spectrums calculated from the in-situ monitoring signals were used as input features for the training. To form the ensemble deep learner, the ensemble weights for the based learners were optimized by differential evolution (DE) (Storn and Price 1997) with the objective of average training accuracy and selected by a designed selection strategy with three constraints including: (1) prediction accuracy, (2) diversity and (3) training speed.

For vehicle type classification, Liu et al., (Liu et al. 2017b) constructed an ensemble deep learning approach based on three types of ResNet (He et al. 2016) architectures, including ResNet-50, ResNet-101, ResNet-152. With respective initializations pretrained on ImageNet (Jia Deng et al. 2009), the three architectures were independently trained on the same training set which had been augmented with balance sampling to obtain diverse base learners. The majority voting was employed as the ensembling criterion to form the ensemble learner.

For vehicle type classification, Qummar et al., (Qummar et al. 2019) constructed an ensemble deep learning approach based on five deep CNN architectures, including ResNet-50 (He et al. 2016), Inception-v3 (Szegedy et al. 2016), Xception (Chollet 2017), Dense-121 and Dense-169 (Huang et al. 2017b). With respective pretrained initializations, the five architectures were independently trained on the same fundus images of Kaggle (https://www.kaggle.com/c/diabetic-retinopathy-detection/data) to generate five different base learners. To form the ensemble learner, Stacking (Wolpert 1992) was employed as the ensembling criterion.

For the prediction of neuromuscular disorders, Khamparia et al. (Khamparia et al. 2020) also proposed to integrate multiple base learners using deep learning. Bhattacharya coefficient was employed to select top gene features and the obtained coefficient was given as input to generate gene features related to muscular disorder. Based on the extracted features, KNN (Altman 1992), DT (Quinlan 1986), linear discriminant analysis (LDA), quadratic discriminant analysis (QDA), RF (Breiman 2001), variants of SVM (Chang and Lin 2011) was utilized to train multiple base learners. To form the ensemble learner, a five-layer convolutional neural network (5-CNN) was built as an implementation of the meta-learning strategy, which took the predictions of the obtained base learners as inputs and output disease diagnosis.

**4.3.3 Summary**

The information of the reviewed recent works of UEDL is listed in Table. 9. In addition to Table. 4, the feature extraction methods adopted by the UDEL approaches proposed in the reviewed works are listed in Table. 10. In addition to Table. 5 and Table. 6, the learning algorithms and learning strategies employed to generate base learners are respectively listed in Table. 11 and Table. 12. In addition to Table. 7, the ensembling criteria utilized to integrate base learners are listed in Table. 13. Referring to Table. 9-13, the key innovations of the UEDL approaches proposed in the reviewed works are summarized in Table. 14. Finally, referring to the paradigm of UEDL presented in Fig. 5, the basic patterns of applying deep learning to TEL to evolve the UEDL approaches proposed in the reviewed works are listed in Table. 15.

Table. 9. The information of the applications in the reviewed recent UEDL works (DA)

| Code | Description | Citations | Published Year |
|---|---|---|---|
| DA1 | Probabilistic wind power forecasting (Wang et al. 2017) | 348 | 2017 |
| DA2 | Electricity load demand forecasting (Qiu et al. 2017) | 229 | 2017 |
| DA3 | Sentiment analysis in social applications (Araque et al. 2017) | 299 | 2017 |
| DA4 | Melanoma recognition in dermoscopic images (Codella et al. 2017) | 295 | 2017 |
| DA5 | Wind speed forecasting (Chen et al. 2018) | 144 | 2018 |
| DA6 | Crude oil price forecasting (Zhao et al. 2017) | 162 | 2017 |
| DA7 | Cancer prediction based on gene expression (Xiao et al. 2018) | 176 | 2018 |
| DA8 | Prediction of melt index in industrial polymerization processes (Liu et al. 2018) | 113 | 2018 |
| DA9 | Brain disease diagnosis based on MRI image (Suk et al. 2017) | 148 | 2017 |
| DA10 | Red lesion detection in fundus images (Orlando et al. 2018) | 117 | 2018 |
| DA11 | Bearing fault diagnosis (Xu et al. 2019) | 65 | 2019 |
| DA11 | Bearing fault diagnosis (Ma and Chu 2019) | 49 | 2019 |
| DA12 | Vehicle type classification (Liu et al. 2017b) | 52 | 2017 |
| DA13 | Diabetic retinopathy detection (Qummar et al. 2019) | 50 | 2019 |
| DA14 | Prediction of neuromuscular disorders (Khamparia et al. 2020) | 45 | 2020 |

Table. 10. In addition to Table. 4, the methods for feature extraction (FE) of the UEDL (TFE/DFE) approaches proposed in the reviewed works

| Code | Description |
|---|---|
| TFE22 | Mallat wavelet decomposition (Mallat 1989) |
| TFE23 | Empirical mode decomposition (EMD) (Huang et al. 1998) |
| TFE24 | Color histogram (Barata et al. 2014) |
| TFE25 | Edge histogram (Barata et al. 2014) |
| TFE26 | Multi-scale variant of color local binary patters (MSLBP) (Zhu et al. 2010) |
| TFE27 | Sparse coding (SC) (Mairal 2014) |
| TFE28 | Raw wind speed series (Chen et al. 2018) |
| TFE29 | Factors including price, flow, stock, macroeconomic and financial series (Zagaglia 2010; Naser 2016) |
| TFE30 | DEseq (Anders and Huber 2010) |
| TFE31 | Volumetric features extracted from MRI image (Suk et al. 2017) |
| TFE32 | Volumetric features extracted from fundus images (Niemeijer et al. 2005; SB and Singh 2012; Seoud et al. 2016) |
| TFE33 | Morlet wavelet transform (LIN and QU 2000) |
| TFE34 | Frequency spectrums calculated from the in-situ monitoring signals (Ma and Chu 2019) |
| TFE35 | Raw fundus images |
| TFE36 | Raw vehicle images |
| TFE37 | gene features related to muscular disorder (Qummar et al. 2019) |
| DFE1 | Word2vec (Mikolov et al. 2013) |

| Code | Description |
| --- | --- |
| DFE2 | Doc2vec (Le and Mikolov 2014) |
| DFE3 | Caffe (Jia et al. 2014) pretrained on ImageNet (Deng et al. 2009) |
| DFE4 | ResNet (He et al. 2016) pretrained on ImageNet (Deng et al. 2009) |
| DFE5 | U-Net (Ronneberger et al. 2015) pretrained on lesion segmentation task (Codella et al. 2018) |
| DFE6 | Unsupervised learning features from chosen factors via DBN (Hinton and Salakhutdinov 2006; Hinton 2012) |
| DFE7 | Supervised learning features via convolutional neural network (Orlando et al. 2018) |

Table. 11. In addition to Table. 5, the learning algorithms (LA) of the UEDL (TLA/DLA) approaches proposed in the reviewed works

| Code | Abbr. | Description |
| --- | --- | --- |
| TLA17 | DT | Decision tree (Quinlan 1986) |
| TLA18 | GBDT | Gradient boosting decision tree (Friedman 2001) |
| TLA19 | KLSM | kernel learning soft-sensing model (Liu et al. 2013) |
| TLA20 | SRM | Sparse regression model (Wang et al. 2011; Zhang and Shen 2012) |
| TLA21 | LDA | Linear discriminant analysis |
| TLA22 | QDA | Quadratic discriminant analysis |
| DLA1 | DCNN | Deep convolutional neural network (LeCun et al. 1998) |
| DLA2 | DBN | Deep belief network (Hinton et al. 2006) |
| DLA3 | LSTM | Long short-term memory (Hochreiter and Schmidhuber 1997; Sainath et al. 2015) |
| DLA4 | SDAE | Stacked denoising autoencoder (Vincent et al. 2008, 2010) |
| DLA5 | ResNet | Deep residual network (He et al. 2016) |
| DLA6 | DAE | Deep auto-encoder (Jia et al. 2016) |
| DLA7 | Inception-v3 | Rethinking the Inception Architecture for Computer Vision (Szegedy et al. 2016) |
| DLA8 | Xception | Deep learning with depthwise separable convolutions (Chollet 2017) |
| DLA9 | Dense | Densely Connected Convolutional Networks (Huang et al. 2017b) |

Table. 12. In addition to Table. 6, the learning strategies (LS) of the UEDL approaches proposed in the reviewed works

| Code | Description |
| --- | --- |
| LS13 | Training with differently initialized values of regularization parameters (Suk et al. 2017) |
| LS14 | Multi-objective evolutionary algorithm (MOEA/DD) (Li et al. 2015) |

Table. 13. In addition to Table. 7, the ensembling criteria (EC) of the UEDL (TEC/DEC) approaches proposed in the reviewed works

| Code | Description |
| --- | --- |
| TEC10 | Wavelet reconstruction (Daubechies and Bates 1993) |
| TEC11 | Random forests (Araque et al. 2017) |
| TEC12 | Support vector regression machine optimized by extremal optimization (Chen et al. 2018) |
| TEC13 | Winner-take-all (Guo et al. 2016) |
| TEC14 | Differential evolution (Storn and Price 1997) |
| TEC15 | Ensemble selection with constraints (Ma and Chu 2019) |

| | | | |
|---|---|---|---|
| DEC1 | Five-layer neural network (Xiao et al. 2018) | | |
| DEC2 | Six-layer convolutional neural network (Suk et al. 2017) | | |
| DEC3 | Five-layer convolutional neural network (Khamparia et al. 2020) | | |

Table. 14. The key innovations of the UEDL approaches proposed in the reviewed works

| App. | FE | LA&S | | EC |
|---|---|---|---|---|
| | | LA | LS | |
| DA1 | TFE22 | DLA1 | Hetero. by LS5 | W. by TEC10 |
| DA2 | TFE23 | DLA2 | Homo. by LS5 | W. by TEC1,8 |
| DA3 | DFE1,2 | TLA8 | Homo. by LS4 | M. by TEC11 |
| DA4 | TFE24-27, DFE3-5 | TLA1 | Homo. by LS5 | W. by TEC8 |
| DA5 | TFE28 | DLA3 | Hetero. by LS4 | M. by TEC12 |
| DA6 | TFE29 | DLA4 | Homo. by LS7 | W. by TEC8 |
| DA7 | TFE30 | TLA1,3,7,17,18 | Homo. by LS4 | M. by DEC1 |
| DA8 | DFE6 | TLA19 | Homo. by LS7 | W. by TEC8 |
| DA9 | TFE31 | TLA20 | Homo. by LS13 | M. by DEC2 |
| DA10 | TFE32, DFE7 | TLA17 | Homo. by LS3 | W. by TEC2 |
| DA11 | TFE33, DFE7 | TLA3 | Homo. by LS3 | W. by TEC13 |
| | TFE34 | DLA2,5,6 | Hetero. by LS14 | W. by TEC14 / S. by TEC15 |
| DA12 | TFE35 | DLA5 | Hetero. by LS4 | W. by TEC2 |
| DA13 | TFE36 | DLA5,7-9 | Hetero. by LS4 | W. by TEC2 |
| DA14 | TFE37 | TLA7,17,21,22 | Hetero. by LS4 | M. by DEC3 |

Table. 15. The basic patterns of applying deep learning to TEL to evolve the UEDL approaches proposed in the reviewed works

| App.1-5 | Pattern | App.6-10 | Pattern | App.11-14 | Pattern |
|---|---|---|---|---|---|
| DA1 | B | DA6 | B | DA11 | A |
| DA2 | B | DA7 | C | | B |
| DA3 | A | DA8 | B | DA12 | B |
| DA4 | A | DA9 | C | DA13 | B |
| DA5 | B | DA10 | A | DA14 | C |

Based on these tables, we can summarize: Five of the reviewed works evolved UEDL approaches by employing pattern A, which utilized pretrained or unsupervised trained deep models to extract more expressive features from raw data; Eight of the reviewed works evolved UEDL approaches by employing pattern B, which replaced the traditional learning algorithms with deep learning algorithms to generate base deep learners with hand-crafted features or raw data; Three of the reviewed works evolved UEDL approaches by employing pattern C, which constructed meta-learning based ensembling criteria to form deep ensemble learner by deep learning algorithms; And one work of DA11 proposed an ensemble selection (S.) based ensembling criterion to reduce the costs of ensemble deep learner at testing, as it was more expensive than traditional ensemble learner. These summarizations reflect the primary underlying trends for the recent development of UEDL.

## 4.4 Unattainability

Based on the primary underlying trends of UEDL summarized in subsection 4.3.3, in this subsection, we discuss the unattainability of UEDL. Although a few learning strategies and ensemble criteria that are more appropriate for the methodology of UEDL have been proposed, the primary issue associated with the methodology of UEDL is that it still retains the paradigm of TEL by simply introducing DL to TEL. This nature of UEDL considerably increases the time and space demands of training multiple base deep learners and testing the ensemble deep learner. To address this, the concept of knowledge distillation (Frosst and Hinton 2018) has become popular in many UEDL approaches. The key idea of knowledge distillation is using a student learner, which is often simpler, to distil knowledge of multiple teacher learners selected from a pool of pre-optimized teacher learners which are often more complex (Parisotto et al. 2016; Shen et al. 2019). Though knowledge distillation can reduce the costs at the testing stage of ensemble deep learning, it still requires large extra expenses during the process of training which prevents adequate usage of UEDL in specific fields. Thus, new methodology for fast ensemble deep learning needs to be further studied.

## 5. Fast Ensemble Deep Learning

Retaining the paradigm of TEL, UEDL ignores the intrinsic characteristics of DL. To address this, taking the inherent characteristics of DL into consideration, fast ensemble deep learning (FEDL) emerges, which more intrinsically reduce the time and space overheads of EDL.

## 5.1 Existing FEDL approaches

To reduce the time overhead of EDL in training multiple base deep learners, Huang et al. (Huang et al. 2017a) proposed an FEDL approach named Snapshot, arguing that the local minima found on the optimizing path of stochastic gradient descent (SGD) (Duchi et al. 2010; Kingma and Ba 2015) can benefit ensembles of deep neural networks. Snapshot utilizes the non-convex nature of deep neural networks and the ability of SGD to converge and escape from local minima as needed. By allowing SGD to converge to local minima multiple times along the optimizing path via cyclic learning rates (Loshchilov and Hutter 2017; Smith 2017), instead of training multiple deep learners independently from scratch, they effectively reduced the time cost of obtaining multiple base deep learners. Garipov et al. (Garipov et al. 2018) found that paths with lower loss values existing between the local minima in the loss plane of deep neural networks, and proposed the fast geometric ensembling (FGE) algorithm to find local minima along these paths via cyclic learning rates. As a result, multiple base deep learners can be obtained with less time overhead. Although Snapshot and FGE can effectively reduce the time of training multiple base deep learners, in order to make full use of Snapshot and FGE to obtain better generalization performance, we need to store the obtained multiple base deep learners and average their predictions to form the final ensemble deep learner. This substantially increases the expenses for the testing stage of the final ensemble deep learner. Observing that the local minima found at the end of each learning rate cycle tend to accumulate at the boundary of the low-value area on the loss plane of deep neural networks, the stochastic weight averaging (SWA) algorithm (Izmailov et al. 2018; Maddox et al. 2019) was proposed to average the local minima at the boundary of low-value area to form the ensemble deep learner. It has been proved that SWA can produce ensemble deep learner with better generalization. At the end of each learning rate cycle, SWA saves the weights of the currently optimized deep neural network and averages them with the weights saved at the end of the last learning rate cycle to form the current ensemble deep learner. SWA can obtain multiple base deep learners under the time expense of training one deep learner and the space expense of additional one deep learner, and effectively reduces the time and space cost for the testing stage of the final ensemble deep learner.

## 5.2 Optional methodology of FEDL

Although several FEDL approaches have been proposed, there still lacks a clear definition for FEDL. Alleviating this situation, Yang et al. (Yang et al. 2021) presented an optional definition and inferred evaluations of FEDL based on observations of existing FEDL approaches. To present a clear definition, the problem of FEDL were divided into three procedures, including: A) training a base deep learner; B) starting from the pre-trained base deep learner to find local minima in the entire parameter space to obtain extra base deep learners; C) integrating the obtained multiple base deep learners to form the ensemble deep learner. The methodology of this definition for FEDL is shown in Fig. 7.

With a deep learning algorithm ($DLA = \{dl_0(*; \theta^{dl_0}), dopt_0(*,*; \theta^{dopt_0})\}$) and a learning strategy ($LS1 = \{ls_0(\theta^{ls_0})\}$), the procedure A) of FEDL can be formally expressed as

$$DL_0 = BaseDeepModelDevelopment(D, T, DLA, LS1) = \{dl_0(*; \theta_u^{dl_0})\},$$

$$\theta_u^{dl_0} = deeply\_evolving(D, T, dl_0(*; \theta^{dl_0}), dopt_0(*,*; \theta^{dopt_0}) \mid ls_0(\theta^{ls_0}))$$

$$= arg\, dopt_0\left(dl_0(D; \theta^{dl_0}), T; \theta^{dopt_0} \mid ls_0(\theta^{ls_0})\right).$$
$$\quad\theta^{dl_0}$$

With the pre-trained $DL_0$, $DLA$ and additional learning strategies ($LS2 = \{ls_1(\theta^{ls_1}), \cdots, ls_b(\theta^{ls_b})\}$), the procedure B) of FEDL can be formally expressed as

$$DL_b = ExtraBaseDeepLearnerGeneration(D, T, DL_0, DLA, LS2) =$$
$$\{dl_1(*; \theta_u^{dl_1}), \cdots, dl_b(*; \theta_u^{dl_b})\},$$
$$dl_b(*) = dl_0(*),$$
$$\theta_u^{dl_b} = deeply\_evolving\left(D, T, dl_0(*; \theta^{dl_0} \mid \theta_u^{dl_{b-1}}), dopt_0(*,*; \theta^{dopt_0}) \mid ls_b(\theta^{ls_b})\right)$$
$$= arg\, dopt_0\left(dl_0(D; \theta^{dl_0} \mid \theta_u^{dl_{b-1}}), T; \theta^{dopt_0} \mid ls_b(\theta^{ls_b})\right).$$
$$\quad\theta^{dl_0}$$

Both ensemble criteria $EC = \{l_e(*; \theta^{l_e}), conf(*,*; \theta^{conf})\}$ for TEL and $DEC = \{dl_e(*; \theta^{dl_e}), dconf(*,*; \theta^{dconf})\}$ for UEDL can be leveraged for fast ensemble deep learner formation. In the context of FEDL, these two criteria are also named as ensemble in model space. Another ensemble criterion that is used by FEDL is ensemble in parameter space, which fuses multiple base deep learners into a single deep learner. Formally, let $ECPS = \{fusing(*, \theta^{fusing})\}$ denote the ensemble criterion in parameter space, the procedure C) of FEDL in parameter space can be formally expressed as follows

$$FDL_e = EnsembleDeepLearnerFormation(\{DL_0, DL_b\}, ECPS)$$
$$= \{fdl_e(*; \theta_u^{fdl_e})\},$$
$$fdl_e(*) = dl_0(*),$$
$$\theta_u^{fdl_e} = fusing(\{\theta_u^{dl_0}, \theta_u^{dl_1}, \cdots, \theta_u^{dl_b}\}; \theta^{fusing}).$$

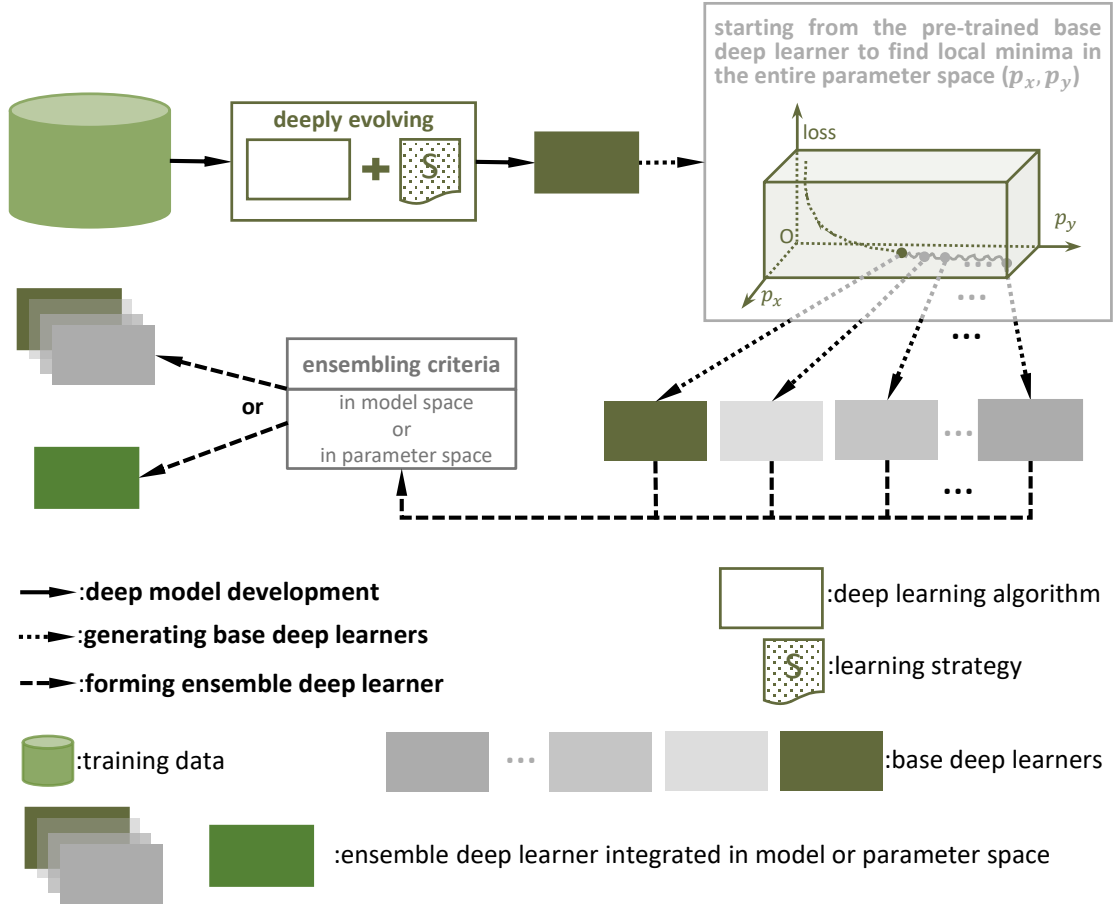

Figure. 7. The methodology of fast ensemble deep learning (FEDL). Deep model development: pre-training a base deep learner. Generating base deep learners: starting from the pre-trained base deep learner to find local minima in the entire parameter space to generate extra base deep learners. Forming ensemble deep learner: integrating the obtained multiple base deep learners to form the final ensemble deep learner.

At testing, given a test data point $d_{test}$, the inference procedures of the ensemble deep learner integrated in model space include computing the predictions of all base deep learners which can be expressed as

$$\tilde{t}_{test} = \{dl_0(d_{test}; \theta_u^{dl_0}), dl_1(d_{test}; \theta_u^{dl_1}), \cdots, dl_b(d_{test}; \theta_u^{dl_b})\},$$

and mapping $\tilde{t}_{test}$ to the final prediction $t_{test}$ by the learner integrated in model space. Whereas, the inference procedure of the ensemble deep learner fused in parameter space can be simply expressed as

$$t_{test} = fdl_e(d_{test}; \theta_u^{fdl_e}).$$

## 5.3 Finding local minima in the sub-parameter space for FEDL

Existing FEDL approaches (like Snapshot, FGE and SWA) have promoted the deployment of ensemble deep learning in some artificial intelligence applications to some extent. However, due to the large expenses compared with TEL, FEDL still needs further advances in some specific fields, where the developing time and computing resources are usually restricted (f. e. the field of robotics vision (Yang et al. 2018, 2019)) or the data to be processed is of large dimensionality (f. e. the field of histopathology whole slide image analysis (Yang et al. 2020c, b)). In these specific fields, it is sometimes still difficult to deploy the TEL approaches, which makes the deployment of existing FEDL approaches still challenging.

To alleviate this, Yang et al (Yang et al. 2020a, 2021) argued that local minima found in the sub-parameter space (LoMiFoSS) can be effective for FEDL, and proposed the concept of LoMiFoSS-based fast ensemble deep learning (LoMiFoSS-FEDL). Referring to the optional methodology presented for FEDL, the problem of LoMiFoSS-FEDL were also divided into three procedures to present the methodology of LoMiFoSS-FEDL, including: A) pre-training a base deep learner; B) starting from the pre-trained base deep learner to find local minima in the sub-parameter space; C) fusing the multiple base deep learners of the found local minima in the sub-parameter space to form the final ensemble deep learner. Different from current state-of-the-art FEDL approaches (Huang et al. 2017a; Garipov et al. 2018; Izmailov et al. 2018; Maddox et al. 2019) that optimize the entire parameters of a deep neural network architecture, LoMiFoSS-FEDL only optimizes partial parameters to further reduce the costs required for training multiple base deep learners. LoMiFoSS-FEDL provides an addition to current FEDL approaches. An optional methodology of LoMiFoSS-FEDL is shown in Fig. 8.

Formally, the procedure A) of LoMiFoSS-FEDL can be expressed exactly as the procedure A) of FEDL. With the pre-trained $DL_0$, $DLA$, a sub-parameter space $SPS = \{\theta^{dl_{0,s}} | \theta^{dl_{0,s}} \in \theta^{dl_0}; \theta^{dl_{0,f}} \in \theta^{dl_0}; (\theta^{dl_{0,s}} \cup \theta^{dl_{0,f}}) == \theta^{dl_0}\}$, and additional learning strategies ($LS2 = \{ls_1(\theta^{ls_1}), \cdots, ls_b(\theta^{ls_b})\}$), the procedure B) of LoMiFoSS-FEDL can be formally expressed as

$$DL_b = ExtraBaseDeepLearnerGeneration(D, T, DL_0, DLA, SPS, LS2) =$$
$$\{dl_1(*; \theta_u^{dl_1}), \cdots, dl_b(*; \theta_u^{dl_b})\},$$
$$dl_b(*) = dl_0(*),$$
$$\theta_u^{dl_b} = \theta_u^{dl_{0,f}} \cup \theta_u^{dl_{b,s}},$$
$$\theta_u^{dl_{b,s}} = partialy\_evolving\left(D, T, dl_0(*; \theta^{dl_{0,s}} | \theta_u^{dl_{b-1}}), dopt_0(*, *; \theta^{dopt_0}) \mid ls_b(\theta^{ls_b})\right)$$
$$= \arg\underset{\theta^{dl_{0,s}}}{dopt_0} \left(dl_0(D; \theta^{dl_{0,s}} | \theta_u^{dl_{b-1}}), T; \theta^{dopt_0} \mid ls_b(\theta^{ls_b})\right).$$

With the ensemble criterion in parameter space $ECPS$, the procedure C) of LoMiFoSS-FEDL in sub-parameter space can be formally expressed as

$$FDL_e = EnsembleDeepLearnerFormation(\{DL_0, DL_b\}, ECPS)$$
$$= \{fdl_e(*; \theta_u^{fdl_e})\},$$
$$fdl_e(*) = dl_0(*),$$

$$\theta_u^{fdl_e} = \theta_u^{dl_{0,f}} \cup \theta_u^{dl_{e,s}},$$
$$\theta_u^{dl_{e,s}} = fusing\left(\left\{\theta_u^{dl_{0,s}}, \theta_u^{dl_{1,s}}, \cdots, \theta_u^{dl_{b,s}}\right\}; \theta^{fusing}\right).$$

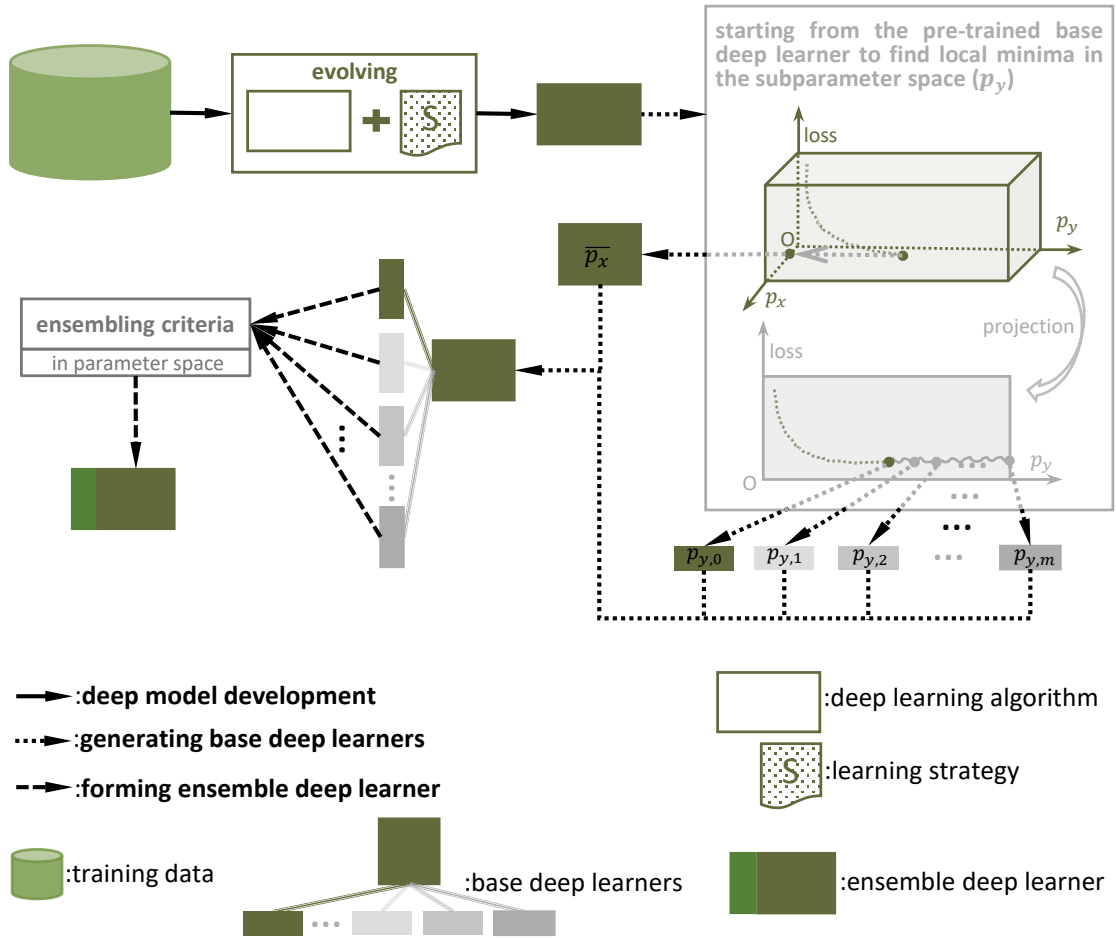

Figure. 8. The methodology of LoMiFoSS-FDEL for fast ensemble deep learning. Deep model development: pre-training a base deep learner. Generating base deep learners: starting from the pre-trained base deep learner to find local minima in the sub-parameter space to generate extra base deep learners. Forming ensemble deep learner: integrating the obtained multiple base deep learners in the sub-parameter space to form the final ensemble deep learner.

At testing, the inference procedure of the ensemble deep learner for LoMiFoSS-FDEL can be expressed the same as the inference procedure of the ensemble deep learner fused in parameter space for FDEL.

### 5.4 Remaining issues

The primary remaining issue associated with FEDL is the imperfection for its definition. From the existing methodologies (Fig. 7 and 8) presented for FEDL, we can notice that the found extra base deep learners are tightly related to the pre-trained base deep learner. However, we cannot be sure whether the found extra base deep learners and the pre-trained base deep learner can be united to produce the optimal ensemble solution. In addition, one may argue that finding multiple deep learners that are less tightly related to each other can achieve better ensemble performances. Thus, the two presented methodologies can only provide some optional perspectives for the definition of FEDL, and new methodologies from different angles for ensemble deep learning are still needed.

## 6 Discussion

This article is different from existing review articles (Sagi and Rokach 2018; Dong et al. 2020) that mostly discussed about traditional ensemble learning, reviews (Cao et al. 2020) that particularly discussed ensemble deep learning in bioinformatics, or specific technical innovations with ensemble learning introduced such as GAN-Ensembles (Durugkar et al. 2017; Ghosh et al. 2018; Han et al. 2021). With more fundamental discussions about the developing routes of both traditional ensemble learning and ensemble deep learning, in this article, we aim to reveal the intrinsic problems and technical challenges of deploying ensemble learning under the era of deep learning to a wider range of specific fields.

Data analyses of published works show that ensemble learning is still prosperously developing while the research of ensemble deep learning (EDL) compared with the research of traditional ensemble learning (TEL) is severely lagging behind. The primary reason for this phenomenon lies in that the time and space overheads for the training and testing stages of EDL are much larger than that of TEL.

Recent advances of TEL have achieved remarkable progresses in various applications. However, most of these advances focused on proposing solutions for specific applications, based on combinatorial innovations by leveraging existing learning strategies for base learner generation and existing ensembling criteria for ensemble learner formation. Few proposed new learning strategies for base learner generation or new ensembling criteria for ensemble learner formation. This phenomenon reflects that the advances for the intrinsic problems of TEL seem to have reached to a bottleneck, although TEL is still prosperously developing in various applications. Besides, the primary issue associated with the methodology of TEL comes from the nature of usual machine learning (UML), which evolves a learner based on hand-crafted features which are usually difficult to design and not expressive enough.

Directly introducing deep learning (DL) to enhance the methodology of TEL by three basic patterns (introducing DL to feature extraction, introducing DL to base learner generation or introducing DL to ensemble learner formation), recent advances of usual EDL (UEDL) have promoted the solutions for various applications. However, the primary issue of the methodology of UEDL is that its nature still retains the paradigm of TEL, which considerably increases the costs of training multiple base deep learners and testing the ensemble deep learner. This can also be confirmed by the fact that knowledge distillation (Frosst and Hinton 2018) and ensemble selection based ensembling criterion (Ma and Chu 2019) for ensemble deep learner formation appeared in some recent advances of UEDL to reduce the testing overhead. A promising future direction is to more intrinsically reduce the expenses of EDL by taking into consideration the inherent characteristics of DL.

Fast ensemble deep learning (FEDL) emerges to more intrinsically reduce the costs of EDL, considering some inherent characteristics of DL. Finding local minima along the (global or local) optimization path of SGD for multiple base deep learners, existing advances of FEDL have provided preferable solutions to promote the deployment of EDL in more artificial intelligence applications of specific fields. However, the primary remaining issue of FEDL is the imperfection for its definition, since currently existing alternative methodologies for FEDL can only provide uncomprehensive perspectives. A promising future direction is to comprehensively propose more appropriate definitions for FEDL.

Notably, many brilliant ideas have been proposed for TEL, another interesting direction is to introduce some ideas of TEL into DL for FEDL, with the reverse direction of the development of UEDL that introduces DL into TEL. For example, recently, Zhang et al (Zhang et al. 2021) introduced negative correlation learning (Liu and Yao 1999) to DL with a divide-and-conquer strategy and proposed a solution of deep negative correlation learning for FEDL.

## Data Availability

Data sharing not applicable to this article as no datasets were generated or analysed during the current study.

## Competing interests

The authors declare that they have no competing interests.

## Reference

Alam M, Samad MD, Vidyaratne L, et al (2020) Survey on Deep Neural Networks in Speech and Vision Systems. Neurocomputing 417:302–321. https://doi.org/10.1016/j.neucom.2020.07.053

Altman NS (1992) An introduction to kernel and nearest-neighbor nonparametric regression. Am Stat. https://doi.org/10.1080/00031305.1992.10475879

Anders S, Huber W (2010) Differential expression analysis for sequence count data. Nat Preced. https://doi.org/10.1038/npre.2010.4282.1

Araque O, Corcuera-Platas I, Sánchez-Rada JF, Iglesias CA (2017) Enhancing deep learning sentiment analysis with ensemble techniques in social applications. Expert Syst Appl. https://doi.org/10.1016/j.eswa.2017.02.002

Bakker B, Heskes T (2003) Clustering ensembles of neural network models. Neural Networks. https://doi.org/10.1016/S0893-6080(02)00187-9

Barata C, Ruela M, Francisco M, et al (2014) Two systems for the detection of melanomas in dermoscopy images using texture and color features. IEEE Syst J. https://doi.org/10.1109/JSYST.2013.2271540

Bauer E, Kohavi R (1999) Empirical comparison of voting classification algorithms: bagging, boosting, and variants. Mach Learn

Behera S, Mohanty MN (2019) Detection of ocular artifacts using bagged tree ensemble model. In: Proceedings - 2019 International Conference on Applied Machine Learning, ICAML 2019

Behler J, Parrinello M (2007) Generalized neural-network representation of high-

dimensional potential-energy surfaces. Phys Rev Lett. https://doi.org/10.1103/PhysRevLett.98.146401

Bi J, Zhang C (2018) An empirical comparison on state-of-the-art multi-class imbalance learning algorithms and a new diversified ensemble learning scheme. Knowledge-Based Syst. https://doi.org/10.1016/j.knosys.2018.05.037

Bishop CM (2006) Pattern Recoginiton and Machine Learning

Boettcher S, Percus A (2000) Nature's way of optimizing. Artif Intell 119:275–286. https://doi.org/10.1016/S0004-3702(00)00007-2

Boettcher S, Percus AG (2001) Optimization with extremal dynamics. Phys Rev Lett. https://doi.org/10.1103/PhysRevLett.86.5211

Boettcher S, Percus AG (2002) Optimization with extremal dynamics. Complexity. https://doi.org/10.1002/cplx.10072

Borş AG, Pitas I (1996) Median radial basis function neural network. IEEE Trans Neural Networks. https://doi.org/10.1109/72.548164

Breiman L (1996) Bagging predictors. Mach Learn. https://doi.org/10.1007/bf00058655

Breiman L (2001) Random forests. Mach Learn. https://doi.org/10.1023/A:1010933404324

Cao Y, Geddes TA, Yang JYH, Yang P (2020) Ensemble deep learning in bioinformatics. Nat. Mach. Intell.

Chang CC, Lin CJ (2011) LIBSVM: A Library for support vector machines. ACM Trans Intell Syst Technol. https://doi.org/10.1145/1961189.1961199

Chen J, Zeng GQ, Zhou W, et al (2018) Wind speed forecasting using nonlinear-learning ensemble of deep learning time series prediction and extremal optimization. Energy Convers Manag. https://doi.org/10.1016/j.enconman.2018.03.098

Chen W, Feng P, Ding H, Lin H (2016) Identifying N 6-methyladenosine sites in the Arabidopsis thaliana transcriptome. Mol Genet Genomics. https://doi.org/10.1007/s00438-016-1243-7

Chen W, Feng PM, Lin H, Chou KC (2013) IRSpot-PseDNC: Identify recombination spots with pseudo dinucleotide composition. Nucleic Acids Res. https://doi.org/10.1093/nar/gks1450

Chen W, Pourghasemi HR, Kornejady A, Zhang N (2017) Landslide spatial modeling: Introducing new ensembles of ANN, MaxEnt, and SVM machine learning techniques. Geoderma. https://doi.org/10.1016/j.geoderma.2017.06.020

Chollet F (2017) Xception: Deep learning with depthwise separable convolutions. In: Proceedings - 30th IEEE Conference on Computer Vision and Pattern Recognition, CVPR 2017

Codella NCF, Gutman D, Celebi ME, et al (2018) Skin lesion analysis toward melanoma detection: A challenge at the 2017 International symposium on biomedical imaging (ISBI), hosted by the international skin imaging collaboration (ISIC). In: 2018 IEEE 15th International Symposium on Biomedical Imaging (ISBI 2018). IEEE, pp 168–172

Codella NCF, Nguyen QB, Pankanti S, et al (2017) Deep learning ensembles for melanoma recognition in dermoscopy images. IBM J Res Dev.

https://doi.org/10.1147/JRD.2017.2708299

Cox DR (1959) The Regression Analysis of Binary Sequences. J R Stat Soc Ser B. https://doi.org/10.1111/j.2517-6161.1959.tb00334.x

Da Silva NFF, Hruschka ER, Hruschka ER (2014) Tweet sentiment analysis with classifier ensembles. Decis Support Syst. https://doi.org/10.1016/j.dss.2014.07.003

Das A, Mohapatra SK, Mohanty MN (2022) Design of deep ensemble classifier with fuzzy decision method for biomedical image classification. Appl Soft Comput 115:108178. https://doi.org/10.1016/J.ASOC.2021.108178

Daubechies I, Bates BJ (1993) Ten Lectures on Wavelets. J Acoust Soc Am 93:1671–1671. https://doi.org/10.1121/1.406784

Davies MN, Secker A, Freitas AA, et al (2008) Optimizing amino acid groupings for GPCR classification. Bioinformatics. https://doi.org/10.1093/bioinformatics/btn382

Deng J, Dong W, Socher R, et al (2009) ImageNet: A large-scale hierarchical image database. In: 2009 IEEE Conference on Computer Vision and Pattern Recognition. IEEE, pp 248–255

Dietterich TG (2000) Ensemble methods in machine learning. In: Lecture Notes in Computer Science (including subseries Lecture Notes in Artificial Intelligence and Lecture Notes in Bioinformatics)

Dietterich TG (1997) Machine-learning research: Four current directions. AI Mag

Dietterich TG, Bakiri G (1995) Solving Multiclass Learning Problems via Error-Correcting Output Codes. J Artif Intell Res. https://doi.org/10.1613/jair.105

Ding S, Zhao H, Zhang Y, et al (2015) Extreme learning machine: algorithm, theory and applications. Artif Intell Rev. https://doi.org/10.1007/s10462-013-9405-z

Dong X, Yu Z, Cao W, et al (2020) A survey on ensemble learning. Front. Comput. Sci.

Dos Santos EM, Sabourin R, Maupin P (2008) A dynamic overproduce-and-choose strategy for the selection of classifier ensembles. Pattern Recognit. https://doi.org/10.1016/j.patcog.2008.03.027

Dou J, Yunus AP, Bui DT, et al (2020) Improved landslide assessment using support vector machine with bagging, boosting, and stacking ensemble machine learning framework in a mountainous watershed, Japan. Landslides 17:641–658. https://doi.org/10.1007/s10346-019-01286-5

Duchi J, Hazan E, Singer Y (2010) Adaptive subgradient methods for online learning and stochastic optimization. In: COLT 2010 - The 23rd Conference on Learning Theory

Durugkar I, Gemp I, Mahadevan S (2017) Generative multi-adversarial networks. In: 5th International Conference on Learning Representations, ICLR 2017 - Conference Track Proceedings

Freund Y, Schapire RE (1997) A Decision-Theoretic Generalization of On-Line Learning and an Application to Boosting. J Comput Syst Sci. https://doi.org/10.1006/jcss.1997.1504

Freund Y, Schapire RE (1996) Experiments with a New Boosting Algorithm. Proc 13th Int Conf Mach Learn. https://doi.org/10.1.1.133.1040

Friedman JH (2001) Greedy function approximation: A gradient boosting machine. Ann

Stat. https://doi.org/10.1214/aos/1013203451

Frosst N, Hinton G rey (2018) Distilling a neural network into a soft decision tree. In: CEUR Workshop Proceedings

Gardner MW, Dorling SR (1998) Artificial neural networks (the multilayer perceptron) - a review of applications in the atmospheric sciences. Atmos Environ. https://doi.org/10.1016/S1352-2310(97)00447-0

Garipov T, Izmailov P, Podoprikhin D, et al (2018) Loss surfaces, mode connectivity, and fast ensembling of DNNs. In: Advances in Neural Information Processing Systems

Geladi P, Kowalski BR (1986) Partial least-squares regression: a tutorial. Anal Chim Acta. https://doi.org/10.1016/0003-2670(86)80028-9

Ghosh A, Kulharia V, Namboodiri V, et al (2018) Multi-agent Diverse Generative Adversarial Networks. In: Proceedings of the IEEE Computer Society Conference on Computer Vision and Pattern Recognition

Granitto PM, Verdes PF, Ceccatto HA (2005) Neural network ensembles: Evaluation of aggregation algorithms. Artif Intell. https://doi.org/10.1016/j.artint.2004.09.006

Guo C, Yang Y, Pan H, et al (2016) Fault analysis of High Speed Train with DBN hierarchical ensemble. In: Proceedings of the International Joint Conference on Neural Networks

Guo Y, Yu L, Wen Z, Li M (2008) Using support vector machine combined with auto covariance to predict protein-protein interactions from protein sequences. Nucleic Acids Res. https://doi.org/10.1093/nar/gkn159

Gupta S, Dennis J, Thurman RE, et al (2008) Predicting human nucleosome occupancy from primary sequence. PLoS Comput Biol. https://doi.org/10.1371/journal.pcbi.1000134

Guyon I, Elisseeff A (2006) Feature Extraction, Foundations and Applications: An introduction to feature extraction. Stud Fuzziness Soft Comput

Han X, Chen X, Liu L-P (2021) GAN Ensemble for Anomaly Detection. Proc AAAI Conf Artif Intell 35:4090–4097. https://doi.org/10.1609/aaai.v35i5.16530

Hansen LK, Salamon P (1990) Neural Network Ensembles. IEEE Trans Pattern Anal Mach Intell. https://doi.org/10.1109/34.58871

He K, Zhang X, Ren S, Sun J (2016) Deep Residual Learning for Image Recognition. In: 2016 IEEE Conference on Computer Vision and Pattern Recognition (CVPR). IEEE, pp 770–778

Hernández-Lobato D, Martinez-Muñoz G, Suárez A (2009) Statistical instance-based pruning in ensembles of independent classifiers. IEEE Trans Pattern Anal Mach Intell. https://doi.org/10.1109/TPAMI.2008.204

Hinton GE (2012) A Practical Guide to Training Restricted Boltzmann Machines. In: Computer. pp 599–619

Hinton GE, Osindero S, Teh YW (2006) A fast learning algorithm for deep belief nets. Neural Comput. https://doi.org/10.1162/neco.2006.18.7.1527

Hinton GE, Salakhutdinov RR (2006) Reducing the dimensionality of data with neural networks. Science (80- ). https://doi.org/10.1126/science.1127647

Ho TK (1998) The random subspace method for constructing decision forests. IEEE

Trans Pattern Anal Mach Intell. https://doi.org/10.1109/34.709601

Hochreiter S, Schmidhuber J (1997) Long Short-Term Memory. Neural Comput. https://doi.org/10.1162/neco.1997.9.8.1735

Hoeting JA, Madigan D, Raftery AE, Volinsky CT (1999) Bayesian model averaging: A tutorial. Stat Sci. https://doi.org/10.1214/ss/1009212519

Hu LY, Huang MW, Ke SW, Tsai CF (2016) The distance function effect on k-nearest neighbor classification for medical datasets. Springerplus. https://doi.org/10.1186/s40064-016-2941-7

Huang G, Huang G Bin, Song S, You K (2015) Trends in extreme learning machines: A review. Neural Networks

Huang G, Li Y, Pleiss G, et al (2017a) Snapshot Ensembles: Train 1, Get M for Free. In: International Conference on Learning Representations 2017

Huang G, Liu Z, Maaten L van der, Weinberger KQ (2017b) Densely Connected Convolutional Networks. In: 2017 IEEE Conference on Computer Vision and Pattern Recognition (CVPR). IEEE, pp 2261–2269

Huang NE, Shen Z, Long SR, et al (1998) The empirical mode decomposition and the Hubert spectrum for nonlinear and non-stationary time series analysis. Proc R Soc A Math Phys Eng Sci. https://doi.org/10.1098/rspa.1998.0193

Izmailov P, Podoprikhin D, Garipov T, et al (2018) Averaging weights leads to wider optima and better generalization. In: 34th Conference on Uncertainty in Artificial Intelligence 2018, UAI 2018

Jia Deng, Wei Dong, Socher R, et al (2009) ImageNet: A large-scale hierarchical image database. In: 2009 IEEE Conference on Computer Vision and Pattern Recognition. IEEE, pp 248–255

Jia F, Lei Y, Lin J, et al (2016) Deep neural networks: A promising tool for fault characteristic mining and intelligent diagnosis of rotating machinery with massive data. Mech Syst Signal Process. https://doi.org/10.1016/j.ymssp.2015.10.025

Jia Y, Shelhamer E, Donahue J, et al (2014) Caffe: Convolutional architecture for fast feature embedding. In: MM 2014 - Proceedings of the 2014 ACM Conference on Multimedia

Jungnickel D (1999) The Greedy Algorithm. pp 129–153

Kang S, Cho S, Kang P (2015) Constructing a multi-class classifier using one-against-one approach with different binary classifiers. Neurocomputing. https://doi.org/10.1016/j.neucom.2014.08.006

Khamparia A, Singh A, Anand D, et al (2020) A novel deep learning-based multi-model ensemble method for the prediction of neuromuscular disorders. Neural Comput Appl. https://doi.org/10.1007/s00521-018-3896-0

Khan A, Sohail A, Zahoora U, Qureshi AS (2020) A survey of the recent architectures of deep convolutional neural networks. Artif Intell Rev 53:5455–5516. https://doi.org/10.1007/s10462-020-09825-6

Kingma DP, Ba JL (2015) Adam: A method for stochastic gradient descent. ICLR Int Conf Learn Represent

Kumar PR (2010) Dynamic programming. In: The Control Systems Handbook: Control System Advanced Methods, Second Edition

Le Q, Mikolov T (2014) Distributed representations of sentences and documents. In: 31st International Conference on Machine Learning, ICML 2014

LeCun Y, Bengio Y, Hinton G (2015) Deep learning. Nature 521:436–444. https://doi.org/10.1038/nature14539

LeCun Y, Bottou L, Bengio Y, Haffner P (1998) Gradient-based learning applied to document recognition. Proc IEEE. https://doi.org/10.1109/5.726791

Lee D, Karchin R, Beer MA (2011) Discriminative prediction of mammalian enhancers from DNA sequence. Genome Res. https://doi.org/10.1101/gr.121905.111

Li K, Deb K, Zhang Q, Kwong S (2015) An evolutionary many-objective optimization algorithm based on dominance and decomposition. IEEE Trans Evol Comput. https://doi.org/10.1109/TEVC.2014.2373386

Li T, Qian Z, He T (2020) Short-Term Load Forecasting with Improved CEEMDAN and GWO-Based Multiple Kernel ELM. Complexity. https://doi.org/10.1155/2020/1209547

Li W, Du Q, Zhang F, Hu W (2016) Hyperspectral Image Classification by Fusing Collaborative and Sparse Representations. IEEE J Sel Top Appl Earth Obs Remote Sens. https://doi.org/10.1109/JSTARS.2016.2542113

LIN J, QU L (2000) FEATURE EXTRACTION BASED ON MORLET WAVELET AND ITS APPLICATION FOR MECHANICAL FAULT DIAGNOSIS. J Sound Vib 234:135–148. https://doi.org/10.1006/jsvi.2000.2864

Liu B, Long R, Chou KC (2016a) IDHS-EL: Identifying DNase i hypersensitive sites by fusing three different modes of pseudo nucleotide composition into an ensemble learning framework. Bioinformatics. https://doi.org/10.1093/bioinformatics/btw186

Liu W, Wang Z, Liu X, et al (2017a) A survey of deep neural network architectures and their applications. Neurocomputing 234:11–26. https://doi.org/10.1016/j.neucom.2016.12.038

Liu W, Zhang M, Luo Z, Cai Y (2017b) An Ensemble Deep Learning Method for Vehicle Type Classification on Visual Traffic Surveillance Sensors. IEEE Access 5:24417–24425. https://doi.org/10.1109/ACCESS.2017.2766203

Liu Y, Gao Z, Chen J (2013) Development of soft-sensors for online quality prediction of sequential-reactor-multi-grade industrial processes. Chem Eng Sci. https://doi.org/10.1016/j.ces.2013.07.002

Liu Y, Yang C, Gao Z, Yao Y (2018) Ensemble deep kernel learning with application to quality prediction in industrial polymerization processes. Chemom Intell Lab Syst. https://doi.org/10.1016/j.chemolab.2018.01.008

Liu Y, Yao X (1999) Ensemble learning via negative correlation. Neural Networks. https://doi.org/10.1016/S0893-6080(99)00073-8

Liu Z, Xiao X, Yu DJ, et al (2016b) pRNAm-PC: Predicting N6-methyladenosine sites in RNA sequences via physical-chemical properties. Anal Biochem. https://doi.org/10.1016/j.ab.2015.12.017

Loshchilov I, Hutter F (2017) SGDR: Stochastic gradient descent with warm restarts. In: 5th International Conference on Learning Representations, ICLR 2017 - Conference Track Proceedings

Ma S, Chu F (2019) Ensemble deep learning-based fault diagnosis of rotor bearing systems. Comput Ind. https://doi.org/10.1016/j.compind.2018.12.012

Maddox W, Garipov T, Izmailov P, et al (2019) A Simple Baseline for Bayesian Uncertainty in Deep Learning. Adv Neural Inf Process Syst

Mairal J (2014) Sparse Modeling for Image and Vision Processing

Mallat SG (1989) A Theory for Multiresolution Signal Decomposition: The Wavelet Representation. IEEE Trans Pattern Anal Mach Intell. https://doi.org/10.1109/34.192463

Mao S, Jiao L, Xiong L, et al (2015) Weighted classifier ensemble based on quadratic form. Pattern Recognit. https://doi.org/10.1016/j.patcog.2014.10.017

Martinez-Muñoz G, Hernández-Lobato D, Suarez A (2009) An analysis of ensemble pruning techniques based on ordered aggregation. IEEE Trans Pattern Anal Mach Intell. https://doi.org/10.1109/TPAMI.2008.78

Masoudnia S, Ebrahimpour R (2014) Mixture of experts: A literature survey. Artif Intell Rev. https://doi.org/10.1007/s10462-012-9338-y

Mendes-Moreira J, Soares C, Jorge AM, De Sousa JF (2012) Ensemble approaches for regression: A survey. ACM Comput. Surv.

Mikolov T, Chen K, Corrado G, Dean J (2013) Efficient estimation of word representations in vector space. In: 1st International Conference on Learning Representations, ICLR 2013 - Workshop Track Proceedings

Mohammed A, Kora R (2021) An effective ensemble deep learning framework for text classification. J King Saud Univ - Comput Inf Sci. https://doi.org/10.1016/j.jksuci.2021.11.001

Mohapatra SK, Khilar R, Das A, Mohanty MN (2021) Design of Gradient Boosting Ensemble Classifier with Variation of Learning Rate for Automated Cardiac Data Classification. In: 2021 8th International Conference on Signal Processing and Integrated Networks (SPIN). IEEE, pp 11–14

Naser H (2016) Estimating and forecasting the real prices of crude oil: A data rich model using a dynamic model averaging (DMA) approach. Energy Econ. https://doi.org/10.1016/j.eneco.2016.02.017

Niemeijer M, Van Ginneken B, Staal J, et al (2005) Automatic detection of red lesions in digital color fundus photographs. IEEE Trans Med Imaging. https://doi.org/10.1109/TMI.2005.843738

Omari A, Figueiras-Vidal AR (2015) Post-aggregation of classifier ensembles. Inf Fusion. https://doi.org/10.1016/j.inffus.2015.01.003

Orlando JI, Prokofyeva E, del Fresno M, Blaschko MB (2018) An ensemble deep learning based approach for red lesion detection in fundus images. Comput Methods Programs Biomed. https://doi.org/10.1016/j.cmpb.2017.10.017

Page L, Brin S, Motwani R, Winograd T (1998) The PageRank Citation Ranking: Bringing Order to the Web. World Wide Web Internet Web Inf Syst. https://doi.org/10.1.1.31.1768

Panda S, Das A, Mishra S, Mohanty MN (2021) Epileptic Seizure Detection using Deep Ensemble Network with Empirical Wavelet Transform. Meas Sci Rev 21:110–116. https://doi.org/10.2478/msr-2021-0016

Parisotto E, Ba J, Salakhutdinov R (2016) Actor-mimic deep multitask and transfer reinforcement learning. In: 4th International Conference on Learning Representations, ICLR 2016 - Conference Track Proceedings

Pham BT, Tien Bui D, Prakash I, Dholakia MB (2017) Hybrid integration of Multilayer Perceptron Neural Networks and machine learning ensembles for landslide susceptibility assessment at Himalayan area (India) using GIS. Catena. https://doi.org/10.1016/j.catena.2016.09.007

Phillips SJ, Dudík M, Schapire RE (2004) A maximum entropy approach to species distribution modeling. In: Proceedings, Twenty-First International Conference on Machine Learning, ICML 2004

Prasad R, Deo RC, Li Y, Maraseni T (2018) Soil moisture forecasting by a hybrid machine learning technique: ELM integrated with ensemble empirical mode decomposition. Geoderma. https://doi.org/10.1016/j.geoderma.2018.05.035

Qiu X, Ren Y, Suganthan PN, Amaratunga GAJ (2017) Empirical Mode Decomposition based ensemble deep learning for load demand time series forecasting. Appl Soft Comput J. https://doi.org/10.1016/j.asoc.2017.01.015

Quinlan JR (1986) Induction of decision trees. Mach Learn. https://doi.org/10.1007/bf00116251

Qummar S, Khan FG, Shah S, et al (2019) A Deep Learning Ensemble Approach for Diabetic Retinopathy Detection. IEEE Access. https://doi.org/10.1109/ACCESS.2019.2947484

Rahmati O, Pourghasemi HR, Zeinivand H (2016) Flood susceptibility mapping using frequency ratio and weights-of-evidence models in the Golastan Province, Iran. Geocarto Int. https://doi.org/10.1080/10106049.2015.1041559

Ramsey FL (1974) Characterization of the Partial Autocorrelation Function. Ann Stat 2:. https://doi.org/10.1214/aos/1176342881

Rawat W, Wang Z (2017) Deep convolutional neural networks for image classification: A comprehensive review. Neural Comput.

Rodríguez JJ, Kuncheva LI, Alonso CJ (2006) Rotation forest: A New classifier ensemble method. IEEE Trans Pattern Anal Mach Intell. https://doi.org/10.1109/TPAMI.2006.211

Ronneberger O, Fischer P, Brox T (2015) U-Net: Convolutional Networks for Biomedical Image Segmentation. In: Lecture Notes in Computer Science (including subseries Lecture Notes in Artificial Intelligence and Lecture Notes in Bioinformatics). pp 234–241

Rosenblatt (1958) The Perceptron: A Theory of Statistical Separability in Cognitive Systems

Rumelhart DE, Hinton GE, Williams RJ (1986) Learning representations by back-propagating errors. Nature. https://doi.org/10.1038/323533a0

Sagi O, Rokach L (2018) Ensemble learning: A survey. Wiley Interdiscip. Rev. Data Min. Knowl. Discov.

Sainath TN, Vinyals O, Senior A, Sak H (2015) Convolutional, Long Short-Term Memory, fully connected Deep Neural Networks. In: 2015 IEEE International Conference on Acoustics, Speech and Signal Processing (ICASSP). IEEE, pp 4580–4584

Salzberg SL (1994) C4.5: Programs for Machine Learning by J. Ross Quinlan. Morgan Kaufmann Publishers, Inc., 1993. Mach Learn. https://doi.org/10.1007/bf00993309

Sandler M, Howard A, Zhu M, et al (2018) MobileNetV2: Inverted Residuals and Linear Bottlenecks. In: Proceedings of the IEEE Computer Society Conference on Computer Vision and Pattern Recognition

SB S, Singh V (2012) Automatic Detection of Diabetic Retinopathy in Non-dilated RGB Retinal Fundus Images. Int J Comput Appl. https://doi.org/10.5120/7297-0511

Schapire RE (1990) The Strength of Weak Learnability. Mach Learn. https://doi.org/10.1023/A:1022648800760

Seoud L, Hurtut T, Chelbi J, et al (2016) Red Lesion Detection Using Dynamic Shape Features for Diabetic Retinopathy Screening. IEEE Trans Med Imaging. https://doi.org/10.1109/TMI.2015.2509785

Shahabi H, Shirzadi A, Ghaderi K, et al (2020) Flood detection and susceptibility mapping using Sentinel-1 remote sensing data and a machine learning approach: Hybrid intelligence of bagging ensemble based on K-Nearest Neighbor classifier. Remote Sens. https://doi.org/10.3390/rs12020266

Shakeel PM, Tolba A, Al-Makhadmeh Z, Jaber MM (2020) Automatic detection of lung cancer from biomedical data set using discrete AdaBoost optimized ensemble learning generalized neural networks. Neural Comput Appl. https://doi.org/10.1007/s00521-018-03972-2

Shen J, Zhang J, Luo X, et al (2007) Predicting protein-protein interactions based only on sequences information. Proc Natl Acad Sci U S A. https://doi.org/10.1073/pnas.0607879104

Shen Z, He Z, Xue X (2019) MEAL: Multi-Model ensemble via adversarial learning. In: 33rd AAAI Conference on Artificial Intelligence, AAAI 2019, 31st Innovative Applications of Artificial Intelligence Conference, IAAI 2019 and the 9th AAAI Symposium on Educational Advances in Artificial Intelligence, EAAI 2019

Smith LN (2017) Cyclical learning rates for training neural networks. In: Proceedings - 2017 IEEE Winter Conference on Applications of Computer Vision, WACV 2017

Storn R, Price K (1997) Differential Evolution - A Simple and Efficient Heuristic for Global Optimization over Continuous Spaces. J Glob Optim. https://doi.org/10.1023/A:1008202821328

Su H, Yu Y, Du Q, Du P (2020) Ensemble Learning for Hyperspectral Image Classification Using Tangent Collaborative Representation. IEEE Trans Geosci Remote Sens. https://doi.org/10.1109/TGRS.2019.2957135

Su H, Zhao B, Du Q, Sheng Y (2016) Tangent Distance-Based Collaborative Representation for Hyperspectral Image Classification. IEEE Geosci Remote Sens Lett. https://doi.org/10.1109/LGRS.2016.2578038

Suk H Il, Lee SW, Shen D (2017) Deep ensemble learning of sparse regression models for brain disease diagnosis. Med Image Anal. https://doi.org/10.1016/j.media.2017.01.008

Szegedy C, Ioffe S, Vanhoucke V, Alemi AA (2017) Inception-v4, inception-ResNet and the impact of residual connections on learning. In: 31st AAAI Conference on

Artificial Intelligence, AAAI 2017

Szegedy C, Vanhoucke V, Ioffe S, et al (2016) Rethinking the Inception Architecture for Computer Vision. In: 2016 IEEE Conference on Computer Vision and Pattern Recognition (CVPR). IEEE, pp 2818–2826

Tan M, Le Q V. (2021) EfficientNetV2: Smaller Models and Faster Training. https://doi.org/10.48550/arxiv.2104.00298

Ting KM, Witten IH (1997) Stacking bagged and dagged models. Proc of ICML'97

Torres ME, Colominas MA, Schlotthauer G, Flandrin P (2011) A complete ensemble empirical mode decomposition with adaptive noise. In: 2011 IEEE International Conference on Acoustics, Speech and Signal Processing (ICASSP). IEEE, pp 4144–4147

Vesanto J, Alhoniemi E (2000) Clustering of the self-organizing map. IEEE Trans Neural Networks. https://doi.org/10.1109/72.846731

Vincent P, Larochelle H, Bengio Y, Manzagol PA (2008) Extracting and composing robust features with denoising autoencoders. In: Proceedings of the 25th International Conference on Machine Learning

Vincent P, Larochelle H, Lajoie I, et al (2010) Stacked denoising autoencoders: Learning Useful Representations in a Deep Network with a Local Denoising Criterion. J Mach Learn Res

Wang G, Jia R, Liu J, Zhang H (2020) A hybrid wind power forecasting approach based on Bayesian model averaging and ensemble learning. Renew Energy. https://doi.org/10.1016/j.renene.2019.07.166

Wang H, Nie F, Huang H, et al (2011) Identifying AD-sensitive and cognition-relevant imaging biomarkers via joint classification and regression. In: Lecture Notes in Computer Science (including subseries Lecture Notes in Artificial Intelligence and Lecture Notes in Bioinformatics)

Wang H zhi, Li G qiang, Wang G bing, et al (2017) Deep learning based ensemble approach for probabilistic wind power forecasting. Appl Energy. https://doi.org/10.1016/j.apenergy.2016.11.111

Wang J, Liu Z, Wu Y, Yuan J (2014) Learning actionlet ensemble for 3D human action recognition. IEEE Trans Pattern Anal Mach Intell. https://doi.org/10.1109/TPAMI.2013.198

Wasserman L (2000) Bayesian model selection and model averaging. J Math Psychol. https://doi.org/10.1006/jmps.1999.1278

Webb GI (2000) MultiBoosting: a technique for combining boosting and wagging. Mach Learn. https://doi.org/10.1023/A:1007659514849

Wei L, Chen H, Su R (2018) M6APred-EL: A Sequence-Based Predictor for Identifying N6-methyladenosine Sites Using Ensemble Learning. Mol Ther - Nucleic Acids. https://doi.org/10.1016/j.omtn.2018.07.004

Wold S, Esbensen K, Geladi P (1987) Principal component analysis. Chemom Intell Lab Syst 2:37–52. https://doi.org/10.1016/0169-7439(87)80084-9

Wolpert DH (1992) Stacked generalization. Neural Networks. https://doi.org/10.1016/S0893-6080(05)80023-1

Wu X, Kumar V, Ross QJ, et al (2008) Top 10 algorithms in data mining. Knowl Inf Syst.

https://doi.org/10.1007/s10115-007-0114-2

WU Z, HUANG NE (2009) ENSEMBLE EMPIRICAL MODE DECOMPOSITION: A NOISE-ASSISTED DATA ANALYSIS METHOD. Adv Adapt Data Anal 01:1–41. https://doi.org/10.1142/S1793536909000047

Xia J-F, Han K, Huang D-S (2009) Sequence-Based Prediction of Protein-Protein Interactions by Means of Rotation Forest and Autocorrelation Descriptor. Protein Pept Lett. https://doi.org/10.2174/092986610789909403

Xia J, Yokoya N, Iwasaki A (2017) A novel ensemble classifier of hyperspectral and LiDAR data using morphological features. In: ICASSP, IEEE International Conference on Acoustics, Speech and Signal Processing - Proceedings

Xiao Y, Wu J, Lin Z, Zhao X (2018) A deep learning-based multi-model ensemble method for cancer prediction. Comput Methods Programs Biomed 153:1–9. https://doi.org/10.1016/j.cmpb.2017.09.005

Xie S, Girshick R, Dollar P, et al (2017) Aggregated Residual Transformations for Deep Neural Networks. In: 2017 IEEE Conference on Computer Vision and Pattern Recognition (CVPR). IEEE, pp 5987–5995

Xu G, Liu M, Jiang Z, et al (2019) Bearing Fault Diagnosis Method Based on Deep Convolutional Neural Network and Random Forest Ensemble Learning. Sensors 19:1088. https://doi.org/10.3390/s19051088

Xu Y, Yang W, Wang J (2017) Air quality early-warning system for cities in China. Atmos Environ. https://doi.org/10.1016/j.atmosenv.2016.10.046

Yang Y, Chen N, Jiang S (2018) Collaborative strategy for visual object tracking. Multimed Tools Appl 77:7283–7303. https://doi.org/10.1007/s11042-017-4633-x

Yang Y, Lv H, Chen N, et al (2021) Local minima found in the subparameter space can be effective for ensembles of deep convolutional neural networks. Pattern Recognit 109:107582. https://doi.org/10.1016/j.patcog.2020.107582

Yang Y, Lv H, Chen N, et al (2020a) FTBME: feature transferring based multi-model ensemble. Multimed Tools Appl 79:18767–18799. https://doi.org/10.1007/s11042-020-08746-4

Yang Y, Wu Y, Chen N (2019) Explorations on visual localization from active to passive. Multimed Tools Appl 78:2269–2309. https://doi.org/10.1007/s11042-018-6347-0

Yang Y, Yang Y, Chen J, et al (2020b) Handling Noisy Labels via One-Step Abductive Multi-Target Learning: An Application to Helicobacter Pylori Segmentation

Yang Y, Yang Y, Yuan Y, et al (2020c) Detecting helicobacter pylori in whole slide images via weakly supervised multi-task learning. Multimed Tools Appl 79:26787–26815. https://doi.org/10.1007/s11042-020-09185-x

You ZH, Lei YK, Zhu L, et al (2013) Prediction of protein-protein interactions from amino acid sequences with ensemble extreme learning machines and principal component analysis. BMC Bioinformatics. https://doi.org/10.1186/1471-2105-14-S8-S10

Zagaglia P (2010) Macroeconomic factors and oil futures prices: A data-rich model. Energy Econ. https://doi.org/10.1016/j.eneco.2009.11.003

Zhang CX, Zhang JS (2011) A survey of selective ensemble learning algorithms. Jisuanji Xuebao/Chinese J. Comput.

Zhang D, Shen D (2012) Multi-modal multi-task learning for joint prediction of multiple regression and classification variables in Alzheimer's disease. Neuroimage. https://doi.org/10.1016/j.neuroimage.2011.09.069

Zhang L, Shi Z, Cheng MM, et al (2021) Nonlinear Regression via Deep Negative Correlation Learning. IEEE Trans Pattern Anal Mach Intell. https://doi.org/10.1109/TPAMI.2019.2943860

Zhang P, He Z (2015) Using data-driven feature enrichment of text representation and ensemble technique for sentence-level polarity classification. J Inf Sci. https://doi.org/10.1177/0165551515585264

Zhang X, Wang J, Zhang K (2017) Short-term electric load forecasting based on singular spectrum analysis and support vector machine optimized by Cuckoo search algorithm. Electr Power Syst Res. https://doi.org/10.1016/j.epsr.2017.01.035

Zhao Y, Li J, Yu L (2017) A deep learning ensemble approach for crude oil price forecasting. Energy Econ. https://doi.org/10.1016/j.eneco.2017.05.023

Zhou ZH (2012) Ensemble methods: Foundations and algorithms

Zhou ZH (2009) Ensemble Learning. In: Encyclopedia of Biometrics

Zhou ZH, Wu J, Tang W (2002) Ensembling neural networks: Many could be better than all. Artif Intell. https://doi.org/10.1016/S0004-3702(02)00190-X

Zhu C, Bichot CE, Chen L (2010) Multi-scale color local binary patterns for visual object classes recognition. In: Proceedings - International Conference on Pattern Recognition

Zoph B, Vasudevan V, Shlens J, Le Q V. (2018) Learning Transferable Architectures for Scalable Image Recognition. In: 2018 IEEE/CVF Conference on Computer Vision and Pattern Recognition. IEEE, pp 8697–8710